\UseRawInputEncoding
\documentclass[letterpaper, 10 pt, conference]{ieeeconf}  

\IEEEoverridecommandlockouts                              

\usepackage[linesnumbered,ruled,vlined]{algorithm2e}
\makeatletter
\let\NAT@parse\undefined
\makeatother

\usepackage{subfigure}
\usepackage{caption}
\usepackage{graphics} 
\usepackage{epsfig} 
\usepackage{mathptmx} 
\usepackage{times} 
\usepackage{amsmath} 
\usepackage{amssymb}  
\usepackage{float}
\usepackage{mathrsfs} %
\DeclareMathAlphabet{\mathcal}{OMS}{cmsy}{m}{n} %
\usepackage{amsfonts}
\usepackage{mathrsfs}
\usepackage{bm}
\usepackage{cite}
\usepackage{hyperref} %
\usepackage{multirow}
\usepackage{siunitx} 
\usepackage{tikz}
\usepackage[linesnumbered,ruled,vlined]{algorithm2e} 
\usepackage[linewidth=1pt]{mdframed} %
\usepackage[export]{adjustbox} %

\usepackage{booktabs}

\newcommand{\red}{\color{red}}

\hypersetup{
	colorlinks=true,
	linkcolor=cyan, 
	filecolor=blue,      
	urlcolor=black,
	citecolor=black,
}

\title{\LARGE \bf
Agile and Safe Trajectory Planning for Quadruped Navigation with Motion Anisotropy Awareness
} 

\author{Wentao Zhang$^{1}$, Shaohang Xu$^{1,2}$, Peiyuan Cai$^{1}$ and Lijun Zhu$^{1}$
\thanks{$^{1}$School of Artificial Intelligence and Automation, Huazhong University of Science and Technology, China, wentaozhang@hust.edu.cn, shaohangxu@hust.edu.cn, m202173177@hust.edu.cn, ljzhu@hust.edu.cn}%
\thanks{$^{2}$School of Data Science, City University of Hong Kong, HKSAR, shaohanxu2-c@my.cityu.edu.hk}%
}

\begin{document}

\maketitle
\thispagestyle{empty}
\pagestyle{empty}
\begin{abstract} 

Quadruped robots demonstrate robust and agile movements in various terrains; however, their navigation autonomy is still insufficient. One of the challenges is that the motion capabilities of the quadruped robot are anisotropic along different directions, which significantly affects the safety of quadruped robot navigation. This paper proposes a navigation framework that takes into account the motion anisotropy of quadruped robots including  kinodynamic trajectory generation, nonlinear trajectory optimization, and nonlinear model predictive control. In simulation and real robot tests, we demonstrate that our motion-anisotropy-aware navigation framework could: (1) generate more efficient trajectories and realize more agile quadruped navigation; (2) significantly improve the navigation safety in challenging scenarios. The implementation is realized as an open-source package at \href{https://github.com/ZWT006/agile_navigation}{https://github.com/ZWT006/agile\_navigation}.
\end{abstract}
\section{INTRODUCTION}
A Quadruped robot is a bionic machine mimicking the locomotion of quadruped animals in nature. It is capable of omnidirectional movement and traversing complex terrains. Recent advancements in locomotion control for quadruped robots include model predictive control (MPC) \cite{di2018dynamic,kim2019highly,kang2022animal,bjelonic2022offline,xu2023robust} and learning-based approaches \cite{kalakrishnan2011learning,peng2020learning,miki2022learning,wu2023learning,caluwaerts2023barkour}. These works demonstrate the stable and agile locomotion over various terrains where the reference trajectory to be tracked is given by the user command.

The recent advancements have indeed facilitated robots in achieving a high level of locomotion in single-motion modes, encompassing movement speed, angular rate, jumping, and traversing rough terrains. Nonetheless, realizing agile motion akin to animals (involving multiple-motion modes at high speed) in the real world continues to pose a significant challenge for current quadruped robots.

\begin{figure}[htbp]
	\centering
	\subfigure[agile movements]{
		\includegraphics[width = 8.3 cm,frame]{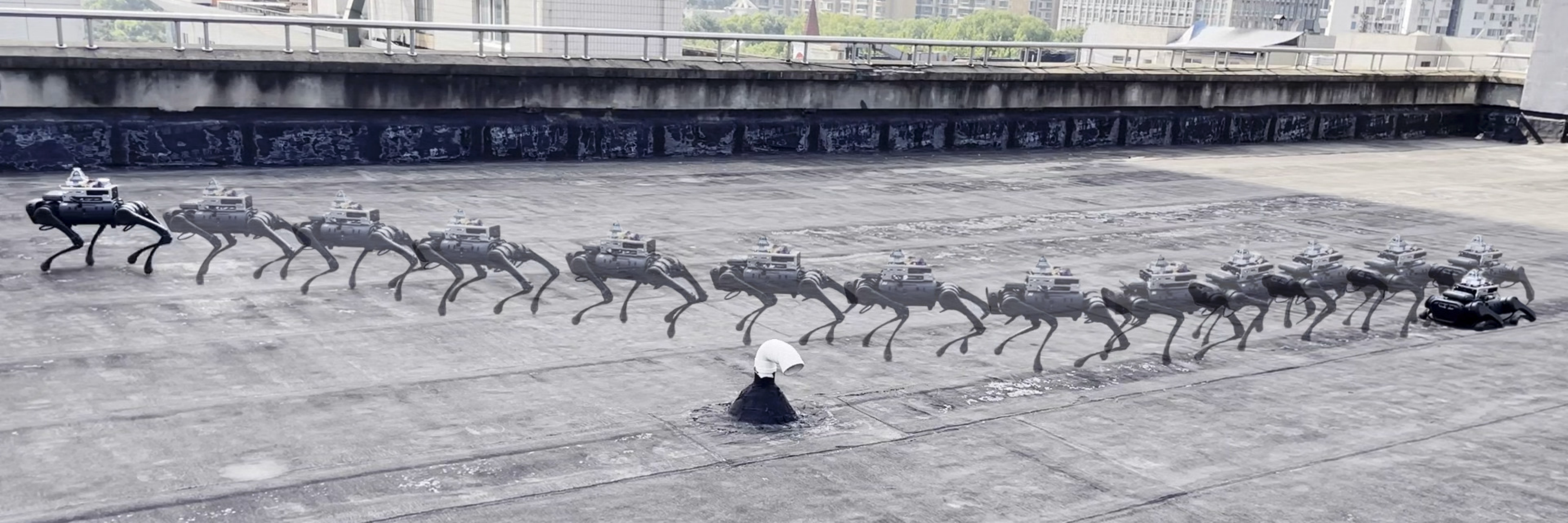}
		\label{fig:Planner}
	}
	\subfigure[planning visualization]{
		\includegraphics[width = 4 cm]{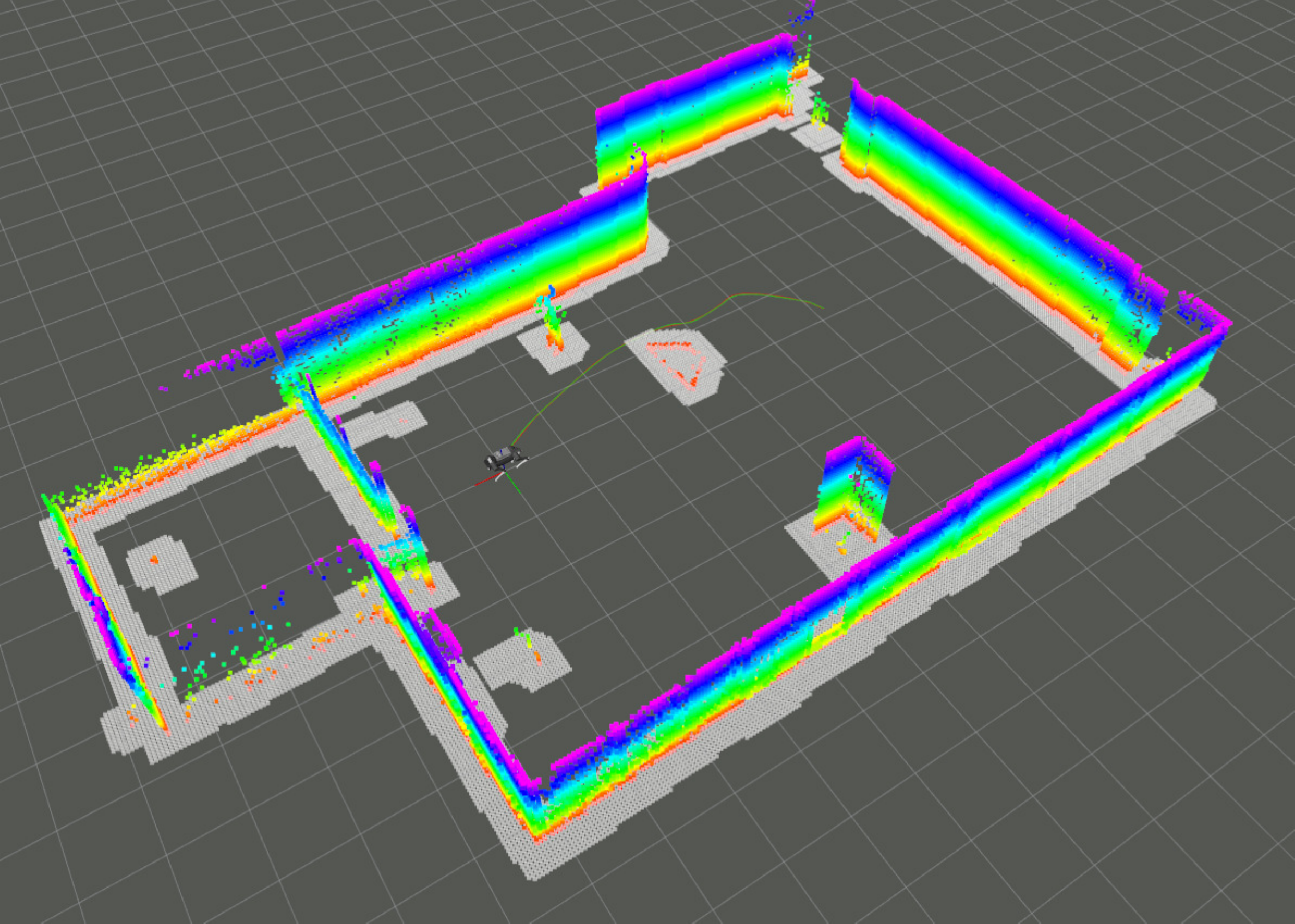}
		\label{fig:RobotSimu} %
	}
	\subfigure[system hardware]{
		\includegraphics[width = 4 cm]{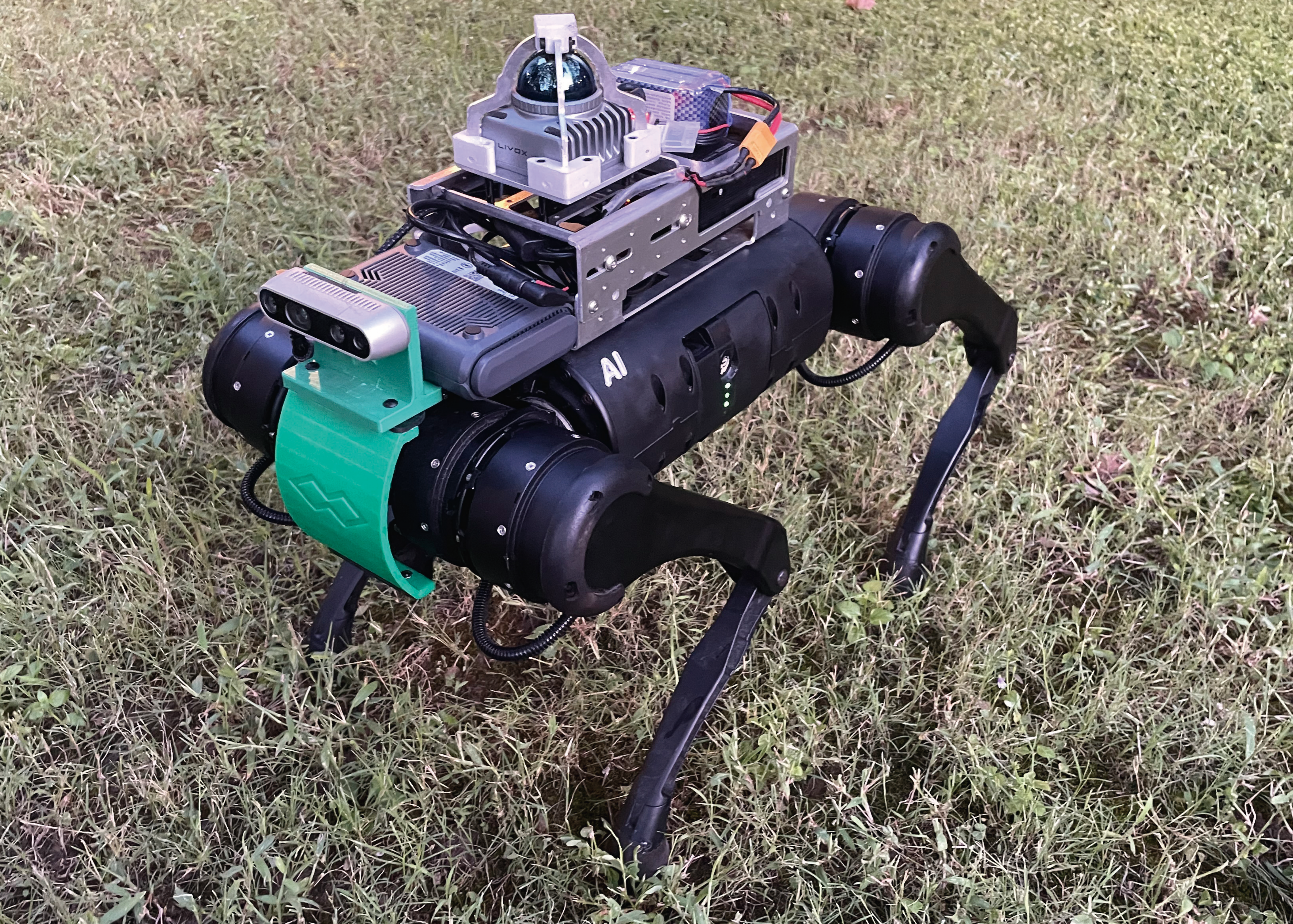}
		\label{fig:RobotReal}
	}
	\caption{(a) showcases our planning system guiding the robot to achieve agile movements in real-world environment. (b) depicts a demonstration of the robot's perception, planning, and control. (c) illustrates the actual hardware deployment of the entire planning system on Unitree A1.}
	\label{fig:highlight}
\end{figure}

Analyzing data from routine motion experiments on quadruped robots, we identify omnidirectional motion anisotropy (OMA) as a key factor influencing their agile motion. This issue is not only observed in the Unitree A1 quadruped robot but is also discussed in \cite{huang2023efficient}, which highlights the motion anisotropy of Cassie biped robot. This anisotropy stems from the mechanical design and joint motors, creating a coupling between the magnitude, direction, and work distance of ground reaction forces (GRF) on the robot's legs and its direction of motion. Therefore, we propose a hierarchical real-time navigation system that takes OMA into account, to achieve agile autonomous quadruped navigation in an unknown/known environment, as Fig. \ref{fig:highlight}. Our experiments demonstrate that neglecting OMA awareness during trajectory planning leads to unstable locomotion when feeding the reference trajectory to the low-level controller.

\subsection{Related Work}

MPC is prevalent technique in legged robot locomotion control, often incorporating collision-free constraints, such as \cite{liao2022walking,gaertner2021collision}. The former is an NMPC-WBC (Whole-Body Control) controller incorporating Discrete Control Barrier Functions, while the latter treats both static and dynamic obstacles as soft inequality constraints. Nevertheless, the effectiveness of control is heavily influenced by the predictive horizon length and the optimization problem is prone to fall into local optima. \cite{zhang2021efficient} focus on person-following navigation in confined spaces through path search, trajectory optimization, and MPC, considering the robot's orientation for collision-free motion planning in geometric space while overlooking posture planning's impact on motion stability. 
Learning-based methods have been proposed as \cite{hoeller2021learning,kim2022learning,wellhausen2021rough,yang2021real}. \cite{hoeller2021learning} demonstrates obstacle avoidance in cluttered and dynamic environments using only depth camera images for learning input. \cite{kim2022learning} proposes a hierarchical navigation framework encompassing mapper, global planner, local planner, and command-tracking controller, yet real-world testing remains pending. Both \cite{wellhausen2021rough} and \cite{yang2021real} leverage neural networks to estimate robots' locomotion capabilities, adapting robot trajectories to dynamics and validating performance on the quadruped robot ANYmal in real-world experiments. However, learning-based approaches require pre-training or dataset. Notably, these datasets are predominantly sourced from simulated environments, potentially introducing discrepancies when transitioning to real-world applications.

Both MPC and RL controllers mentioned above require the operator to provide a trajectory. However, due to the lack of consideration for OMA, following the given trajectory may lead to poor tracking and even result in robot falls (as demonstrated in our subsequent experiments).

Multiple navigation frameworks based on sample-based or search-based path planning methods and optimization methods are proposed in \cite{wermelinger2016navigation,norby2020fast,dudzik2020robust,gilroy2021autonomous,chignoli2022rapid,jelavic2023lstp}. \cite{norby2020fast} considers legged robot flight phases and utilizes Rapidly-Exploring Random Trees (RRT) to generate Center of Mass (COM) trajectories but it has not been tested on real robots. \cite{dudzik2020robust} presents a framework that encompasses visual processing, state estimation, path planning, and locomotion control. Nevertheless, it yields geometrically straight trajectories, lacking smoothness. \cite{gilroy2021autonomous} and \cite{chignoli2022rapid} incorporate jumping into the planning process and have been tested in real world, but with slow movement speeds.

The aforementioned planning approaches do not explicitly consider OMA, which may result in trajectories that are difficult for actual robots to track, such as fast lateral walking. Although \cite{huang2023efficient} considers lateral stability in reactive planning, this reactive method does not take obstacle avoidance into account.

\subsection{Contribution} 
We model the OMA of quadruped robots and introduce a real-time hierarchical navigation framework that specifically addresses the OMA in different level. The key contributions of our study can be outlined as follows:
\begin{itemize}
	\item An efficient kinodynamic trajectory generation method based on LazyPRM* is proposed, which takes the translational motion direction and orientation changes into explicit consideration. The Optimal Boundary Value Problem (OBVP) is employed to kinodynamic connect the state points. 
	\item A spatiotemporal Nonlinear Trajectory Optimization (NTO) problem based on parameterized polynomial trajectory and nonlinear constraint is constructed to further smoothen the trajectory locally and make it dynamically feasible. The OMA is introduced as a elliptical constraint on the maximum translational velocity and motion direction.
	\item An NMPC with WBC locomotion controller is proposed to track the target COM state trajectory, while the penalty on maximum simultaneous linear and angular velocities are introduced to prevent potential falls. 
	\item The proposed agile navigation framework is deployed on the Unitree A1 robot platform in simulations and real-world experiments with unknown environments. We also released it as ros packages, in order to facilitate research communication in relevant fields. 
\end{itemize}

The remainder of this article is organized as follows. The hierarchical planning is elaborated in detail in Section \ref{sec:method}. In Section \ref{sec:experiments}, we integrate perception into the navigation framework and deploy it on the A1 robot platform for real-world implementation. We analyze and evaluate our method through simulations and experiments. Section \ref{sec:conclusion} concludes the paper.
\section{hierarchical planner method}\label{sec:method} 
\subsection{Omnidirectional Motion Anisotropy}
\begin{figure}[htbp]
	\centering
	\subfigure[Leg Joint]{
		\includegraphics[width = 4.7 cm]{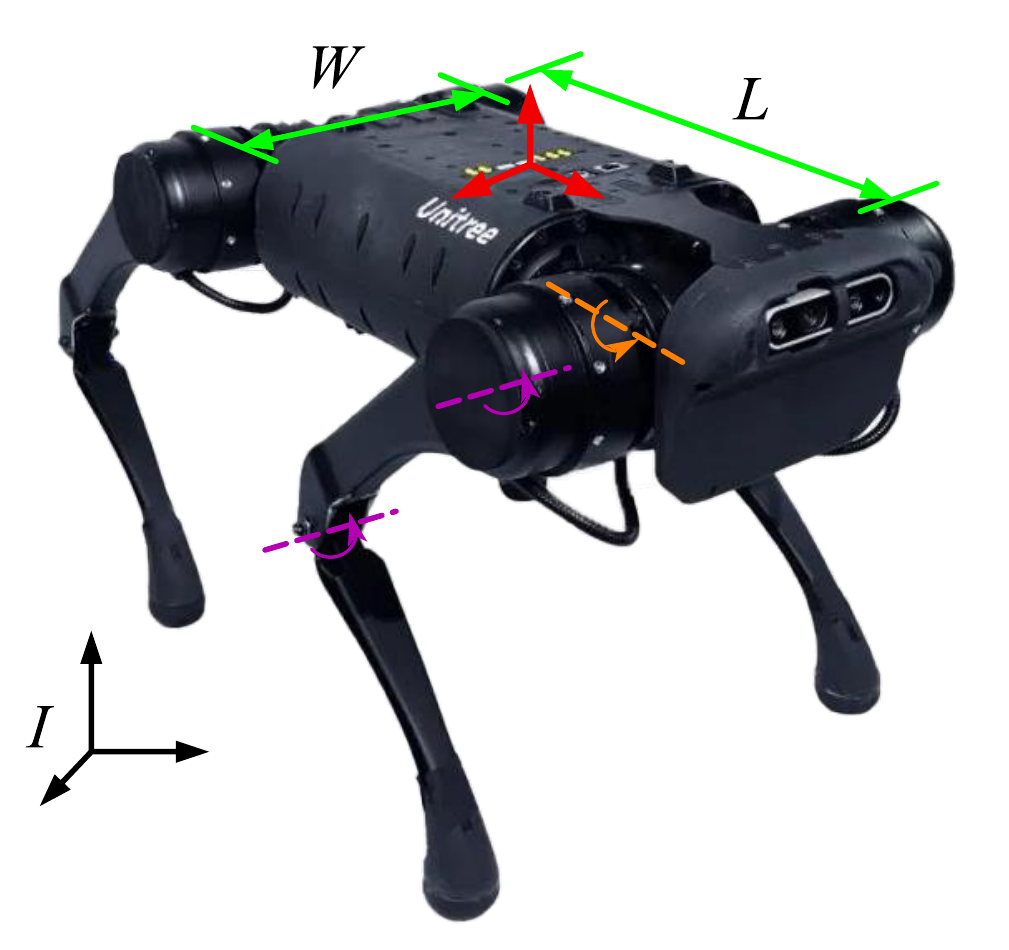}
		\label{fig:legjoint}
	}
	\subfigure[Motion Capability]{
		\includegraphics[width = 3.3 cm]{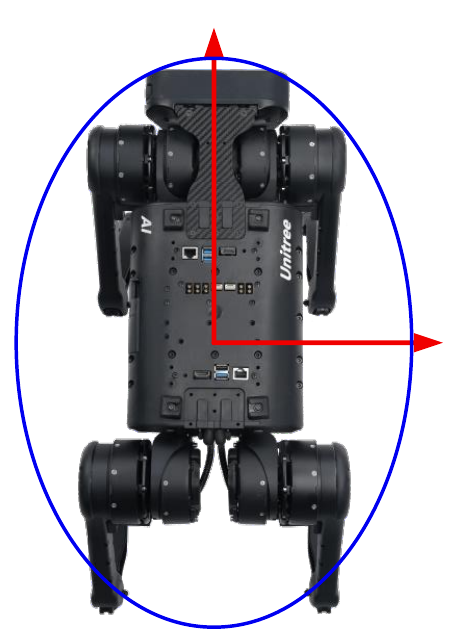}
		\label{fig:OvalAxis}
	}
	\caption{$I$ represents the world inertial coordinate system, and $B$ represent the robot coordinate system. (a) illustrates the drivetrain of a single leg. The orange color represents the shoulder joint, while the purple color indicates the hip and knee joints. The distance between the left and right legs is denoted as $W$, and the distance between the front and back legs is represented by $L$. (b) depicts the robot's motion capabilities in different directions using a blue ellipse. It can be interpreted as the maximum acceleration, maximum velocity, and the farthest distance covered in one gait contact cycle.}
	\label{fig:robotOMA}
\end{figure}

A quadruped robot relies on the GRF generated through leg-ground contact to enable movement. The resultant force of the GRF can act in any direction, allowing the robot to achieve omnidirectional motion. Additionally, both turning and translation motions of the robot depend on the combined forces of GRF, leading to a coupling effect between turning and translation. Consequently, high turning speed and high translation speed cannot be achieved simultaneously. The analysis of OMA in quadruped robot is shown in Fig. \ref{fig:robotOMA}. Each leg consists three joints: the shoulder joint provides lateral movement for roll motion, while the hip and knee joints enable pitch motion for forward and backward locomotion. Furthermore, the distance $L$ is larger than the distance $W$, resulting in unequal work distance in the two directions during one gait contact phase. 

\subsection{Kinodynamic Trajectory Generation} \label{sec:trajgen} 
This algorithmic aims to rapidly generate a coarse robot COM , denoted as $\Gamma$, taking into account both translational and turning movements of the robot by separately planning the plane position $x,y$ and yaw angle $\theta$. Unlike conventional planning methods that consider $\theta =\mathrm{atan}2\left( \dot{x},\dot{y} \right)$ as described in \cite{norby2020fast}, planning $[x,y,\theta]$ separately is crucial to overcome OMA. The algorithm utilizes LazyPRM* \cite{hauser2015lazy}, which is a variant of PRM* \cite{karaman2011sampling}. LazyPRM* checks connectivity only when those two nodes are part of the optimal solution, increasing search efficiency. 

Alg.\ref{alg:kinolazyprm} outlines the key steps. $\mathcal{O}$ and $\mathcal{C}$ refer to the open and closed set of A*\cite{hart1968formal} and $RoadMap$ is the set of state sample points. \textbf{Expend}($\cdot$) searches neighboring nodes $n_c$ around the parent node $n_n$ within a given range, and saves their index sequences in $nearsets$. \textbf{CollisionFree}($\cdot$) checks the safety of trajectory. \textbf{Update}($\cdot$) is normal A* search update process. \textbf{Solve}($\cdot$) is used to solve the OBVP to connect two state points as explained in Section \ref{sec:kinoconnec}.

\IncMargin{1em}
\begin{algorithm}[htbp]
	\SetKwData{PoseMap}{$RoadMap$}	
	\SetKwData{xcur}{$\boldsymbol{p_{\mathrm{cur}}}$}	
	\SetKwData{xpar}{$\boldsymbol{p_{\mathrm{par}}}$}	
	\SetKwData{Nodec}{$n_c$}
	\SetKwData{Noden}{$n_n$}
	\SetKwData{Nodet}{$n_t$}
	\SetKwData{OpenList}{$\mathcal{O}$}	
	\SetKwData{CloseList}{$\mathcal{C}$}	
	\SetKwData{Pose}{$\boldsymbol{p}$}	
	\SetKwData{xstart}{$\boldsymbol{p}_{\mathrm{start}}$}	
	\SetKwData{xgoal}{$\boldsymbol{p}_{\mathrm{goal}}$}		
	\SetKwData{gnCost}{$g_c$}
	\SetKwData{hnCost}{$h_c$}
	\SetKwData{fnCost}{$f_c$}
	\SetKwData{nearsets}{$nearsets$}
	\SetKwData{IDX}{$idx$}
	\SetKwData{IDY}{$idy$}
	\SetKwData{MaxX}{$idx_{\mathrm{max}}$}
	\SetKwData{MaxY}{$idy_{\mathrm{max}}$}
	\SetKwData{GRID}{$\delta _{\mathrm{g}}$}
	\SetKwData{BIAS}{$\delta _{\mathrm{b}}$}

	\SetKwFunction{Sample}{{\bf sample}}	
	\SetKwFunction{Insert}{{\bf insert}}		
	\SetKwFunction{pop}{{\bf pop}}
	\SetKwFunction{Expend}{{\bf Expend}}			
	\SetKwFunction{Update}{{\bf Update}}
	\SetKwFunction{CollisionFree}{{\bf CollisionFree}} 
	\SetKwFunction{ObstacleFree}{{\bf ObstacleFree}}
	\SetKwFunction{Empty}{{\bf empty}}	
	\SetKwFunction{Heuristic}{{\bf Heuristic}}		
	\SetKwFunction{push}{{\bf push}}
	\SetKwFunction{TrajCost}{{\bf EdgeCost}}
	\SetKwFunction{GetPath}{{\bf GetPath}}
	\SetKwFunction{Solve}{{\bf Solve}}
	\SetKwFunction{Contain}{{\bf contain}}

	\SetKwInOut{Input}{Input}	
	\SetKwInOut{Output}{Output}

	\Input{Obstacle map $Map$, Start pose \xstart, Goal pose \xgoal} 
	\Output{A trajectory $\Gamma$ from \xstart to \xgoal}
	\For {\IDX $\leftarrow 0$ \KwTo \MaxX}{
		\For {\IDY $\leftarrow 0$  \KwTo \MaxY}{
			\Pose{$x$} $ = idx\cdot \delta _{\mathrm{g}}+rand\left( -1,1 \right) \cdot \delta _{\mathrm{b}}$\\
			\Pose{$y$} $ = idy\cdot \delta _{\mathrm{g}}+rand\left( -1,1 \right) \cdot \delta _{\mathrm{b}}$\\
			$RoadMap[idx,idy]$ $\leftarrow$ \Pose;\\
		}
	}
	Initialize();\\
	\For{$iter \leftarrow 1$ \KwTo $MaxIter$}{
		\If{\OpenList.\Empty}{
			\Return;\\
		}
		\Noden $ \leftarrow $ \OpenList.\pop{};\CloseList.\Insert{\Noden};\\
		\If{\Noden.\Pose == \xgoal}{
			$\Gamma \leftarrow$ \GetPath{\CloseList,\Noden};\\
			\Return{$\Gamma$};\\
		}
		\xpar $\leftarrow$ \Noden.\Pose;\\
		\nearsets $\leftarrow$ \Expend{\PoseMap,\xpar};\\
		\For{{\IDX.\IDY} $\leftarrow$ \nearsets}{
			\Pose $\leftarrow $ $RoadMap[idx,idy]$;\\
			\If{\Noden.\Pose{$x,y$} == \xgoal{$x,y$}}{
				\Pose{$\theta$} = \xgoal{$\theta$};\\
			}
			\Else {
				\Pose{$\theta$} $ =\mathrm{atan}2 \left( \left( y-y_{\mathrm{par}} \right) /\left( x-x_{\mathrm{par}} \right) \right)$;\\
			}
			\Nodec $\leftarrow $ \Solve{\Pose,\Noden};\\
			\If{\CollisionFree{\Nodec}}
			{
				\Update{\Nodec,\Noden,\CloseList,\OpenList};\\
			}
		}
	}
	\caption{Kinodynamic LazyPRM*}
	\label{alg:kinolazyprm}
\end{algorithm}
\DecMargin{1em}

\subsubsection{RoadMap Sample}
A quadruped robot has 6 Degrees of Freedom (DoF), but only the horizontal position and orientation angle are mainly concerned during the locomotion, since body height, pitch angle, and roll angle are relevant to the terrain. To enhance efficiency, we replaced random sampling in LazyPRM* by uniformly distributed sampling with random perturbation as described in Alg. \ref{alg:kinolazyprm} lines 1 to 5. In this scheme, the entire plane space is uniformly discretized into $idx_{\mathrm{max}}\times idy_{\mathrm{max}} = n$ grids, each grid associated with a state point stored in the $RoadMap$. The state of grid is represented as $RoadMap[idx,idy]=\left[ x,y,\theta \right]$, where $idx$ and $idy$ are the index of grid. $\delta _{\mathrm{g}}$ is grid size and $\delta _{\mathrm{b}}$ is the maximum bias magnitude. Initially, the yaw angle $\theta $ is set to zero and is adjusted during the expanding process based on the states of the parent node and the current node (see Alg. \ref{alg:kinolazyprm} lines 18 to 21). By using the grid indices $idx$ and $idy$ to access state points during the search expansion, the time complexity of finding the optimal trajectory is $\mathrm{O}\left(n\right)$. In contrast, if a completely random sampling approach is used for the entire space and Dijkstra's method is employed to find the optimal solution, the time complexity increases to $\mathrm{O}\left(n^2\right)$. 
\subsubsection{Kinodynamic Connection} \label{sec:kinoconnec}
The motion of quadruped robot is dependent on GRF, which provide acceleration, so we express the simplified motion equation of COM as a double integration system as
\begin{equation*}
	\begin{split}
	\dot{\boldsymbol{s}} &=\mathbf{A}\boldsymbol{s}+\mathbf{B}\boldsymbol{u}\\
	\mathbf{A} &=\left[ \begin{matrix}
		0&		\mathbf{I}_3\\
		0&		0\\
	\end{matrix} \right] ,\mathbf{B}=\left[ \begin{array}{c}
		0\\
		\mathbf{I}_3\\
	\end{array} \right] \\
	\boldsymbol{s}&=\left[ \boldsymbol{p},\boldsymbol{v} \right] ^{\mathrm{T}}=\left[ \boldsymbol{p},\dot{\boldsymbol{p}} \right] ^{\mathrm{T}}\\
	\boldsymbol{u}&=\boldsymbol{a}=\ddot{\boldsymbol{p}}
	\end{split}
\end{equation*}
where $\boldsymbol{s}=\left[ \boldsymbol{p},\boldsymbol{v} \right],\boldsymbol{p}=\left[ p_{\mathrm{x}},p_{\mathrm{y}},p_{\mathrm{\theta}} \right]$ is state variables, system input $\boldsymbol{u}$ is fictitious force that can be achieved by the low-level controller via GRF.

To enhance the consistency between the coarse trajectory and the robot's motion, we formulate the kinodynamic connection between two nodes as an OBVP that minimizes the total trajectory energy $J (T)=\int_0^T{\boldsymbol{a}}(t)^2dt$, with given initial state $\boldsymbol{s}_{\mathrm{i}}=\left[ \boldsymbol{p}_{\mathrm{i}},\boldsymbol{v}_{\mathrm{i}} \right]$, fixed final position state $\boldsymbol{p}_{\mathrm{f}}$, and free final velocity state $\boldsymbol{v}_{\mathrm{f}}$. We employ Pontryagin's maximum principle\cite{mueller2015computationally} to obtain explicit solutions as:
\begin{equation}
	\label{closedformOBVP}
	\begin{split}
		\left[ \begin{array}{c}
			p_{\mu}^{*}\left( t \right)\\
			v_{\mu}^{*}\left( t \right)\\
		\end{array} \right] &=\left[ \begin{array}{c}
			\frac{\alpha _{\mathrm{\mu}}}{6}t^3-\frac{\alpha _{\mathrm{\mu}}T}{2}t^2+v_{\mathrm{\mu i}}t+p_{\mathrm{\mu i}}\\
			\frac{\alpha _{\mathrm{\mu}}}{2}t^2-\alpha _{\mathrm{\mu}}Tt+v_{\mathrm{\mu i}}\\
		\end{array} \right]  \\
		u_{\mu}^{*}\left( t \right) &=\alpha _{\mathrm{\mu}}t-\alpha _{\mathrm{\mu}}T \\
		J^*(T) &=\sum_{\mu \in \{x,y,\theta \}}{\left( \frac{1}{3}\alpha _{\mu}^{2}T^3 \right)} \\
		\alpha _{\mathrm{\mu}} &=-\frac{p_{\mathrm{\mu f}}-v_{\mathrm{\mu i}}t+p_{\mathrm{\mu i}}}{3}.
	\end{split}
\end{equation}
To numerically calculate the solution, we require the value of trajectory duration time $T$, which we specify as $T = T_{\mathrm{ref}}:=\max \left( \left\| \left[ \varDelta x,\varDelta y \right] \right\| _2/v_{\mathrm{ref}},\varDelta \theta /\omega _{\mathrm{ref}} \right)$. $v_{\mathrm{ref}}$ denotes the reference linear velocity and $\omega _{\mathrm{ref}}$ represents the reference angular rate. It is worth noting that the optimal trajectory in this case is a polynomial trajectory.
\subsubsection{Trajectory Cost}
\begin{figure}[htbp] %
	\centering
	\subfigure[]{
		\includegraphics[width=2.5 cm]{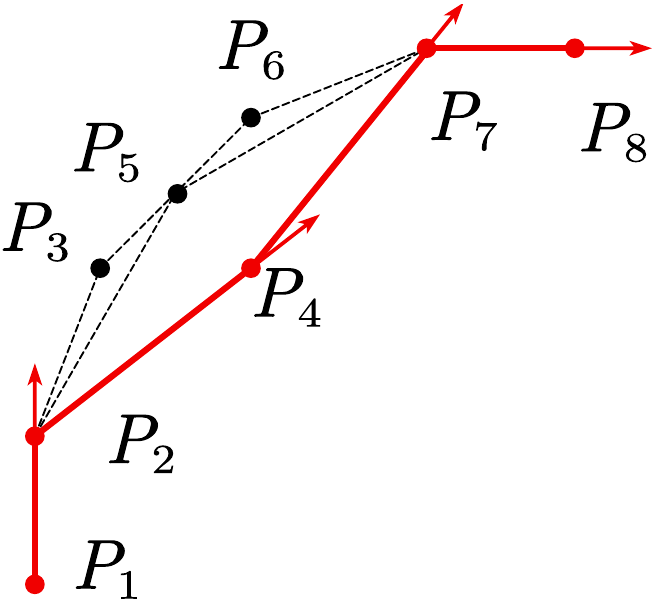}
		\label{AngleCostA}
		}
	\subfigure[]{
		\includegraphics[width=2.5 cm]{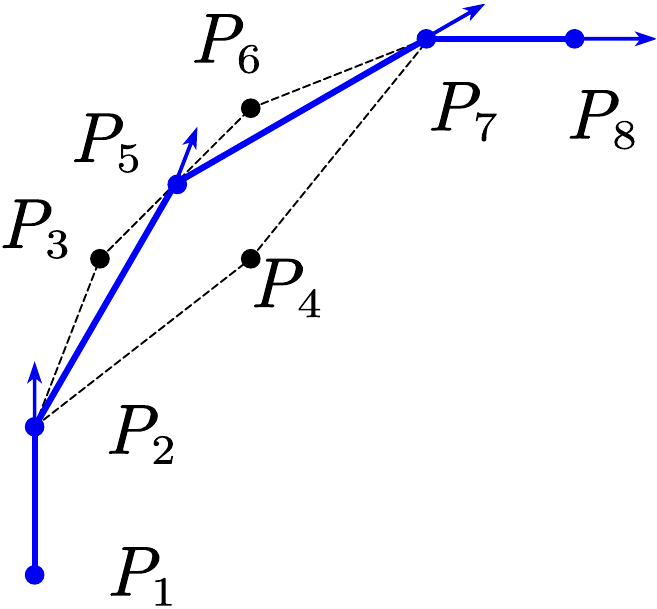}
		\label{AngleCostB}
		}
	\subfigure[]{
		\includegraphics[width=2.5 cm]{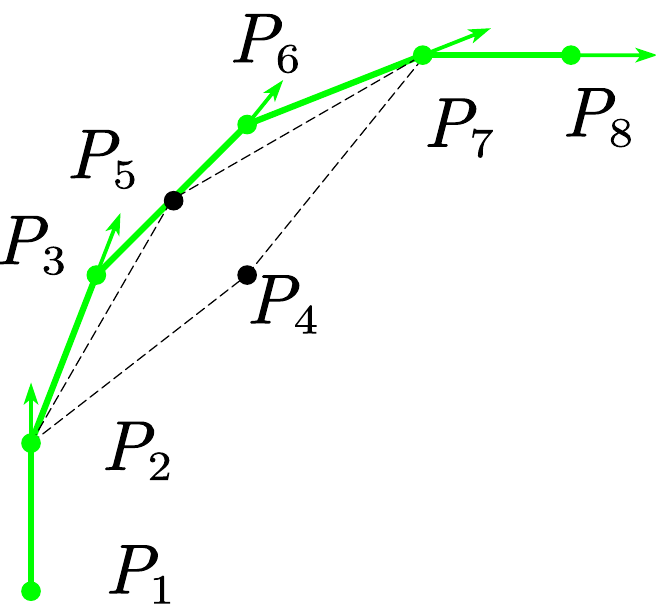}
		\label{AngleCostC}
		}
	\caption{Three trajectories are illustrated connecting $P_1$ and $P_8$. Arrows indicate the yaw angle direction. The translational distance is consistent between (a) and (b), but (b) has a smaller angular variation, demonstrating the effect of introducing an angle cost in trajectory cost. The angular variation is consistent between (b) and (c), but the quadratic angle cost results in a smaller angular cost for (c), demonstrating the effect of using a quadratic angle cost in trajectory cost. In conclusion, \eqref{equ:tc} guides the generation of smooth trajectories.}
	\label{fig:L2normAngleCost}
\end{figure}

The yaw angle and linear velocity direction have a significant impact on the stability of robot motion. Therefore, we propose the following trajectory cost  
\begin{equation} \label{equ:tc}
		t_{\mathrm{c}}=\lambda_{\mathrm{yaw}}\cdot \left | \varDelta \theta \right |^2 +\int_{}^T{\sqrt{\delta x^2+\delta y^2}} 
\end{equation}
where the first term represents yaw cost with weight coefficient $\lambda_{\mathrm{yaw}}$ and the second term is the length of the trajectory. We utilize a quadratic form to promote smoother yaw angle variations. The effect of trajectory cost is shown in Fig. \ref{fig:L2normAngleCost}. The reference linear/angular velocity $v_{\mathrm{ref}}$ and $\omega _{\mathrm{ref}}$ impact the maximum velocity. The reference time $T_{\mathrm{ref}}$ and angle cost weight $\lambda_{\mathrm{yaw}}$ influences trajectory smooth. It is essential to highlight that, unlike many existing approaches, our method specifically incorporates yaw information in the trajectory calculation, enabling the consideration and enforcement of the OMA.

\subsection{Nonlinear Trajectory Optimization}
The rough trajectory $\Gamma$ may be dynamics infeasible, as we using double integrate system to connect waypoints. Moreover, the trajectory duration $T$ in \eqref{closedformOBVP} is determined by the reference velocity, which may not fully exploit the robot's motion capabilities. To address these limitations, we introduce an NTO problem to further improve the quality of $\Gamma$. Taking inspiration from \cite{wang2022geometrically,foehn2021time,zhou2019robust}, we still parameterize trajectory as polynomial but increase its order to improve its flexibility and accuracy. 

\subsubsection{Optimization Problem}  
The optimization problem is formulated with the polynomial parameters $\boldsymbol{c}$ and the duration time $\boldsymbol{T}$ serving as the decision variables, as below: 
\begin{subequations}
	\begin{align}
		\min_{\boldsymbol{c},\boldsymbol{T}} \quad & \sum_{j=1}^N{\left[ \begin{array}{c}
			\lambda _{\mathrm{s}}\int_0^{T_{\mathrm{j}}}{\boldsymbol{u}_{\mathrm{j}}}(\boldsymbol{c}_{\mathrm{j}})^{\mathrm{T}}R\boldsymbol{u}_{\mathrm{j}}(\boldsymbol{c}_{\mathrm{j}})dt+\\
			\lambda _{\mathrm{c}}\phi \left( \boldsymbol{s}_{\mathrm{j}}\left( t \right) \right) +\lambda _{\mathrm{t}}\rho (T_{\mathrm{j}}) \label{tocostfunction}
		\end{array} \right]}
		\\
		\mathrm{s}.\mathrm{t}. \quad & \boldsymbol{s}\left( t \right) = \sum_{i=0}^n{\boldsymbol{c}_{\mathrm{ji}}t^i},i\geqslant 0,\forall t\in [0,T_{\mathrm{j}}],j=0\sim N \label{polynomialst}
		\\
		& G \left( \boldsymbol{s}(t),\dot{\boldsymbol{s}}\left( t \right) \right) \preceq 1 \label{quadrupedst}
		\\
		& H \left(\dot{\boldsymbol{s}}\left( t \right), \ddot{\boldsymbol{s}}\left( t \right) \right) \preceq 0 \label{accelerationst}
		\\
		& \boldsymbol{s}_\mathrm{j-1}^{\left( m \right)}(T_\mathrm{j-1}) = \boldsymbol{s}_\mathrm{j}^{\left( m \right)} (0),m=0\sim M,j=1\sim N \label{continuityst}
	\end{align}	
	\label{equ:not}
\end{subequations}
where $\boldsymbol{s}(t)$ is the $n$th orders polynomial trajectory of $\left[x,y,\theta\right]$, $N$ represents the segment number of trajectories, which is one less than waypoints, $j$ denotes the ${j}$th segment, $R$ is a positive definite diagonal matrix judge states cost, $\boldsymbol{T}$ is the duration time vector, and $\boldsymbol{c}_{\mathrm{ji}}$ represents the ${i}$th order coefficient of the $j$th segment polynomial.

\subsubsection{Objective Function}

The nonlinear optimization problem considers trajectory energy, obstacle avoidance, and time optimality by incorporating cost functions \eqref{tocostfunction} with weights $\lambda_{\mathrm{s}}$, $\lambda_{\mathrm{c}}$, and $\lambda_{\mathrm{t}}$ for each factor, respectively. The $j$th segment trajectory energy cost is described as
\begin{equation}
	\label{statecost}
	\int_0^{T_{\mathrm{j}}}{\boldsymbol{u}_{\mathrm{j}}}(\boldsymbol{c}_\mathrm{j})^{\mathrm{T}} R \boldsymbol{u}_{\mathrm{j}}(\boldsymbol{c}_\mathrm{j})dt.
\end{equation} 
Note that $\boldsymbol{c}_\mathrm{j}=\{\boldsymbol{c}_\mathrm{j1},\cdots,\boldsymbol{c}_\mathrm{jn}\}$, $\boldsymbol{c}=\{\boldsymbol{c}_\mathrm{1},\cdots,\boldsymbol{c}_\mathrm{N}\}$, and the controller input $\boldsymbol{u}_\mathrm{j}$ is written as a function of variables $\boldsymbol{c}_\mathrm{j}$. 

We discretize the spatial trajectory and calculate the obstacle cost to ensure obstacle avoidance as
\begin{equation}\
	\begin{split}
		\phi \left( s_{\mathrm{j}}\left( t \right) \right) &=\int_0^{T_{\mathrm{j}}}{f_{\mathrm{d}}\left( d\left( p\left( t \right) \right) \right) ds} \\
		&\approx\sum_{l=0}^{T_{\mathrm{j}}/\delta t}{f_{\mathrm{d}}\left( d\left( p\left( T_{\mathrm{jl}} \right) \right) \right)} \cdot v  \left( T_{\mathrm{jl}} \right) \cdot \delta t,\;\;T_{\mathrm{jl}}=l\cdot \delta t
	\end{split}
\end{equation}
where $d_{\mathrm{s}}$ is the differentials of the arc lengths of the $\left[ x,y \right]$ trajectories, and $f_{\mathrm{d}}(\cdot)$ is the distance penalty function. An example of such a function is $f_{\mathrm{d}}\left( d\left( p\left( t \right) \right) \right) ={{1}/{d\left( p\left( t \right) \right)}}$. Where $d\left( p\left( t \right) \right)<d_\mathrm{th}$, $d\left( p\left( t \right) \right)$ is the Euclidean Distance Field (EDF) \cite{han2019fiesta} at point $p\left( t \right)$, and $d_{\mathrm{th}}$ is the threshold of obstacle clearance. 

The time penalty function is defined as
\begin{equation}
	\begin{split}
		\label{timecost}
		\rho (T_{\mathrm{j}})&=\left| T_{\mathrm{j}} \right|,T_{\mathrm{j}}>0
		\\
		&=e^{\tau _{\mathrm{j}}},\tau _{\mathrm{j}}\in \mathbb{R}
	\end{split}
\end{equation}
where the time vector $\boldsymbol{T}$ is positive real numbers. we introduce intermediate variable $\boldsymbol{\tau}$ as real number to transform time into an unconstrained optimization variable.

\subsubsection{System Constraints}
The OMA is considered as constraint terms in \eqref{quadrupedst}. As shown in Fig. \ref{fig:OvalAxis}, we use \textbf{elliptical} constraint on the translational maximum velocity as
\begin{equation}  
	\begin{array}{c} \left[ \begin{array}{c}
			v_{\mathrm{Bx}}\\
			v_{\mathrm{By}}\\
		\end{array} \right] =R\left( \theta \right) \left[ \begin{array}{c}
			v_{\mathrm{Ix}}\\
			v_{\mathrm{Iy}}\\
		\end{array} \right]\\
		\frac{v_{Bx}^{2}}{v_{\mathrm{mx}}}+\frac{v_{By}^{2}}{v_{\mathrm{my}}}\leqslant 1
	\end{array}
\label{ellipsevelconstrain}
\end{equation}
where $\theta$ is body yaw angle and $R\left( \theta \right)$ is rotation matrix from world inertial frame $I$ to robot body frame $B$. The maximum frontal and lateral velocity are denoted as $v_{\mathrm{mx}}$ and $v_{\mathrm{my}}$ respectively, with $v_{\mathrm{my}}<v_{\mathrm{mx}}$ indicating that frontal movement capability is stronger than lateral. \eqref{accelerationst} enforces velocity and acceleration limits $\left| \dot{\boldsymbol{s}}\left( t \right) \right|-\boldsymbol{v}_{\max}\leqslant 0,\left| \ddot{\boldsymbol{s}}\left( t \right) \right|-\boldsymbol{a}_{\max}\leqslant 0$. Continuity constraints \eqref{continuityst} ensure smooth transitions between waypoints, where $M$ is the continuous state order.

\begin{figure*}[htbp]
	\centering
	\includegraphics[width = 16 cm]{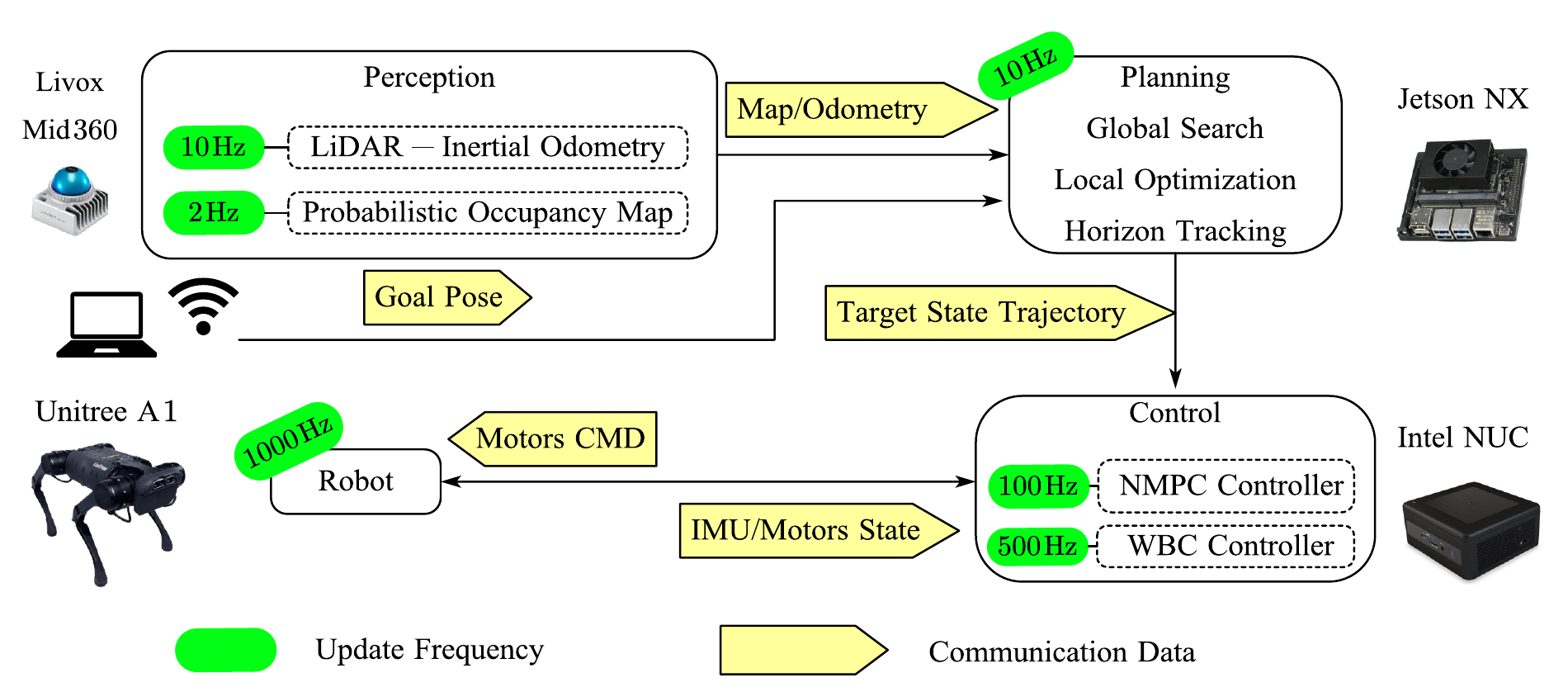}
	\caption{This figure summarizes the proposed planning system. The perception and planning board is Jetson Xavier NX with Graphics Processing Unit (GPU). Hybrid solid-state lidar Livox Mid-360 can provide \SI{360}{\degree} horizontal, \SI{52}{\degree} vertical FOV and \SI{100}{m} detection range. The control board is Intel NUC 11PAHi7-1165G7 with 16G RAM, which runs the locomotion controller NMPC-WBC. Quadruped robot Unitree A1 is the experimental platform. The hardware devices are all connected within the same Local Area Network (LAN) and communicate by the User Datagram Protocol (UDP).}
	\label{fig:systemoverview}
\end{figure*}
\subsection{NMPC Trajectory Tracking} 
We use the optimized trajectory as the reference trajectory in the NMPC framework and introduce velocity safety constraints to ensure the stability during fast motion. 

\subsubsection{NMPC Problem}
The general formulation of NMPC problem is derived from \cite{grandia2023perceptive} as
\begin{subequations}
	\begin{align}
		\min_{\boldsymbol{u}\left( t \right)} \quad & \Phi (\boldsymbol{x}(T))+\int_0^T{L}(\boldsymbol{x}(t),\boldsymbol{u}(t),t)\mathrm{d}t \label{NMPCcost}
		\\
		\mathrm{s}.\mathrm{t}.\quad & \boldsymbol{x}(0)=\hat{\boldsymbol{x}} \label{NMPCinitial} 
		\\
		& \dot{\boldsymbol{x}}=\boldsymbol{f}^c(\boldsymbol{x},\boldsymbol{u},t) \label{NMPCdynamic}
		\\
		& \boldsymbol{g}(\boldsymbol{x},\boldsymbol{u},t)=\boldsymbol{0} \label{NMPCequality}
		\\
		& \boldsymbol{h}(\boldsymbol{x},\boldsymbol{u},t)<\boldsymbol{0} \label{NMPCinequality}
	\end{align}
	\label{NMPCproblem}
\end{subequations} 
where $\boldsymbol{x}\left( t \right)$ and $\boldsymbol{u}\left( t \right)$ are the state and input at time $t$, $\varPhi(\cdot)$ is final state cost and $L(\cdot)$ is quadratic cost function regarding the tracking error of the state and input. $\hat{\boldsymbol{x}}$ represent current measured state. $\boldsymbol{f}^c(\cdot)$ represents system dynamic function with further details provided in \cite{bellicoso2016perception} as the floating base dynamic equation. $\boldsymbol{g}(\cdot)$ and $\boldsymbol{h}(\cdot)$ represent equality constraints and inequality constraints, respectively. The robot state vector is 
$$
\boldsymbol{x}=\left[ \boldsymbol{\theta }_{\mathrm{B}}^{\mathrm{T}},\boldsymbol{p}_{\mathrm{B}}^{T},\boldsymbol{\omega }_{\mathrm{B}}^{T},\boldsymbol{v}_{\mathrm{B}}^{T},\boldsymbol{q}_{\mathrm{j}}^{\mathrm{T}} \right] ^{\mathrm{T}},\boldsymbol{u}=\left[ \boldsymbol{f}_{\mathrm{c}}^{\mathrm{T}},\boldsymbol{v}_{\mathrm{j}}^{\mathrm{T}} \right] ^{\mathrm{T}}
$$
where $B$ represents the robot body frame. In particular, the system state $\boldsymbol{x}$ contains centroidal orientation as Euler angle $\boldsymbol{\theta }_{\mathrm{B}}\in \mathbb{R} ^3$, position $\boldsymbol{p}_{\mathrm{B}}\in \mathbb{R} ^3$, angular rate $\boldsymbol{\omega }_{\mathrm{B}}$, linear velocity $\boldsymbol{v}_{\mathrm{B}}$ and joint position $\boldsymbol{q}_{\mathrm{j}}\in \mathbb{R} ^{12}$. System input $\boldsymbol{u}$ consists of four feet contact forces $\boldsymbol{f}_{\mathrm{c}}\in \mathbb{R} ^4$ and joint velocity $\boldsymbol{v}_{\mathrm{j}}$.

\subsubsection{Safety Constraint} 
During high-speed motion, we introduce a constraint to prevent simultaneous large values of translational and angular velocity. This constraint is imposed to avoid instability and potential falls. Therefore, we incorporate the following constraint:
\begin{equation*} 
	\begin{aligned}
		h_{\mathrm{s}}=\lambda _{\mathrm{l}}\cdot \left( v_{\mathrm{Bx}}^{2}+v_{\mathrm{By}}^{2} \right) +\lambda _{\mathrm{\theta}}\cdot \omega_{\mathrm{B\theta}}^{2}-h_{\mathrm{s}}^{th}<0
	\end{aligned}
\end{equation*}
where the weights $\lambda _{\mathrm{l}}$ and $\lambda _{\mathrm{\theta}}$ govern the trade-off between translational and angular velocity, while $h_{\mathrm{s}}^{th}$ represents safety constraint threshold.

\section{Experiments and Results}\label{sec:experiments}
Our planning method is subjected to numerical computations, physical simulations, and real-world experiments. The search and optimization components are numerical computation tested in MATLAB. Our planning framework is implemented in C++ with ROS as a communication middleware. 

\subsection{Optimization Solving}
For numerical solving of NTO problem, we utilize the general optimization solver NLopt{\footnote{Steven G. Johnson NLopt: \href{http://github.com/stevengj/nlopt}{http://github.com/stevengj/nlopt}}} and employ the gradient-based local optimization method MMA(Method of Moving Asymptotes) \cite{svanberg2002mma}. We utilize the parameters of $\Gamma$ as an initial guess to facilitate solving the NTO problem. In practical testing, it is nearly impossible to solve the problem without an initial guess. We also manually solve the equality constraints during the solving iterative process and replace inequality constraints with penalty function, transforming NOT into an unconstrained optimization problem to improve solving efficiency. This approach enables the application in real-time planning for solving as \SI{10}{Hz}.

\subsection{Planning System Implementation}\label{sec:planningsystem}
The overall navigation system for real-time navigation experiments is illustrated in Fig. \ref{fig:systemoverview}. LiDAR-inertial odometry \cite{xu2022fast} is a fast, robust, and versatile framework, that provides \SI{10}{Hz} high-accuracy odometry and global pointcloud data with lower computation. We use \cite{han2019fiesta} to update the local probabilistic occupancy map centered on the robot at \SI{2}{Hz}, which is not sensitive to dynamic obstacles and reduces the impact of dynamic point clouds on planning. We use OpenCV to calculate 2D EDF map from global occupancy map for path search and trajectory optimization. 

The locomotion controller{\footnote{legged\_control:\href{https://github.com/qiayuanliao/legged\_control}{https://github.com/qiayuanliao/legged\_control}}} is responsible for robot locomotion control and trajectory tracking, and is an NMPC-WBC controller implemented using OCS2 \cite{OCS2}, operates at \SI{100}{Hz} for NMPC and \SI{500}{Hz} for WBC.

\subsection{Simulation} 
We conduct planning simulations in Gazebo, using a corridor obstacle map with size of 30[m]$\times$8[m] and a local square perception radius of \SI{6}{m}. Our planner is set to $v_{\mathrm{ref}}=$\SI{1.5}{m/s} and $\omega_{\mathrm{ref}}=$\SI{1.0}{rad/s} for comparison with champ\cite{CHAMP}, which employs GMapping\cite{GMapping} for SLAM and OMPL\cite{OMPL} for planning. As depicted in Tab. \ref{tab:comparison}, our proposed method achieves significantly higher linear velocity and angular rate, while reducing navigation time by 36\%.
\begin{table}[htbp]
	\renewcommand{\arraystretch}{1.2}
	\centering
	\caption{Simulation Result}\label{tab:comparison}
	\resizebox{8.4cm}{!}{
	\begin{tabular}{cccccc} 
	\hline
	\hline
	    & \multicolumn{5}{c}{Trajectory State}  \\ 
			 \cline{2-6}
	Method  & $time$[\SI{}{s}] & $v_{\mathrm{ave}}$[\SI{}{m/s}] & $\omega_{\mathrm{ave}}$[\SI{}{rad/s}] & $v_{\mathrm{max}}$[\SI{}{m/s}] & $\omega_{\mathrm{max}}$[\SI{}{rad/s}] \\
	\hline
	baseline    	&111.1   &0.2468	&0.0905	&0.6012	&0.7785  \\ 
	proposed      	&\red{40.1}    &\red{1.0024}	&\red{0.5673} &\red{1.6819}	&\red{3.0731}	\\ 
	\hline
	\hline
	\end{tabular}
	}
\end{table}

\subsection{Legged Robot Omnidirectional Constraint}
\begin{figure*}[htbp]
	\centering
	\subfigure[$x$ state trajectory]
	{
		\begin{minipage}[b]{.315\linewidth}
			\centering
			\includegraphics[width=5.6 cm]{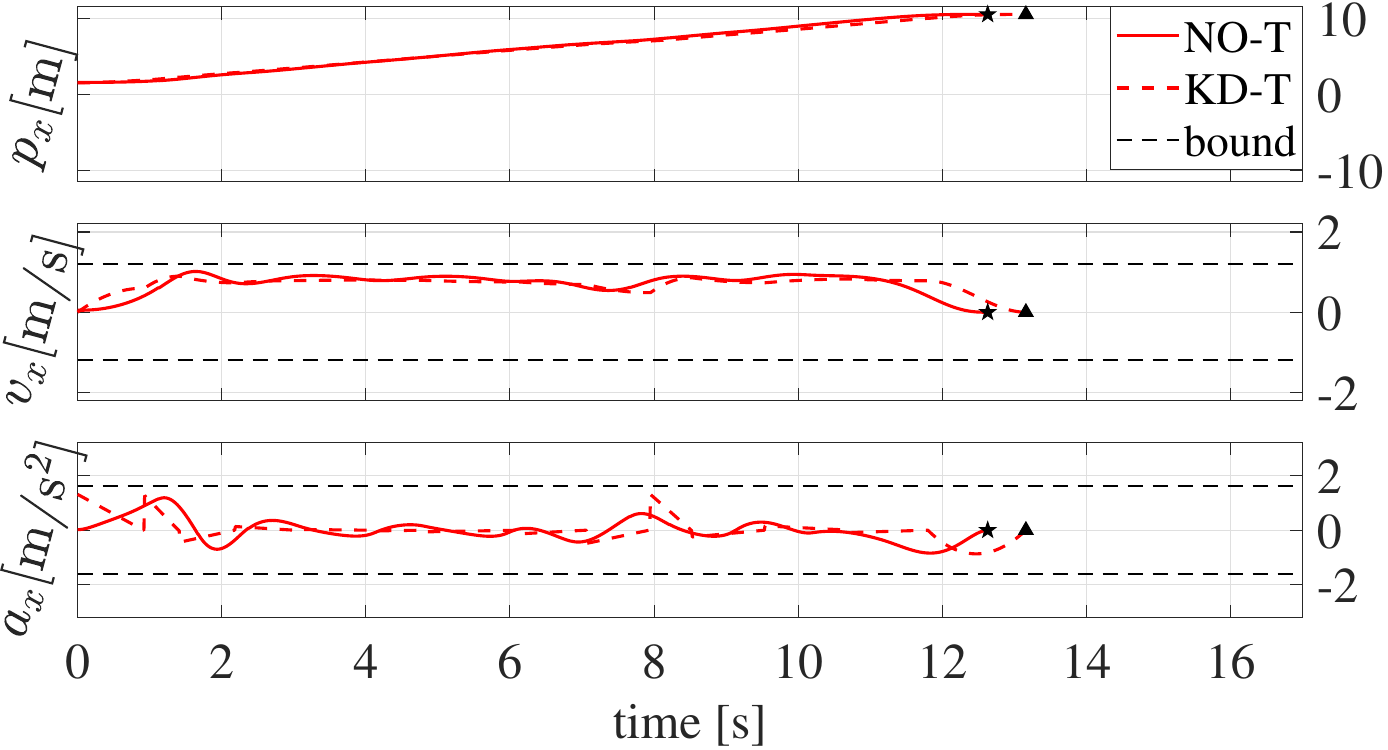}
		\end{minipage}
		\label{fig:xtraj}
	}
	\subfigure[$y$ state trajectory]
	{
		\begin{minipage}[b]{.315\linewidth}
			\centering
			\includegraphics[width=5.6 cm]{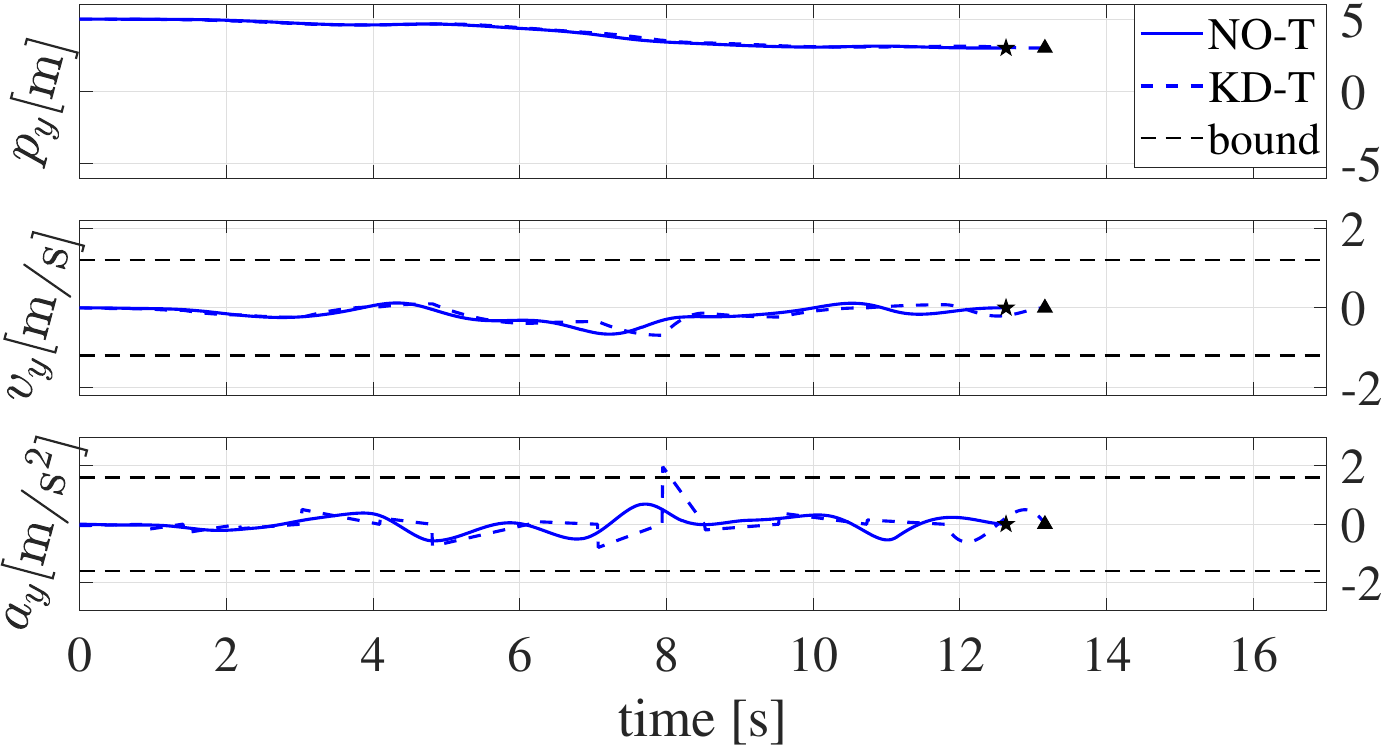}
		\end{minipage}
		\label{fig:ytraj}
	}
	\subfigure[$\theta$ state trajectory]
	{
		\begin{minipage}[b]{.315\linewidth}
			\centering
			\includegraphics[width=5.6 cm]{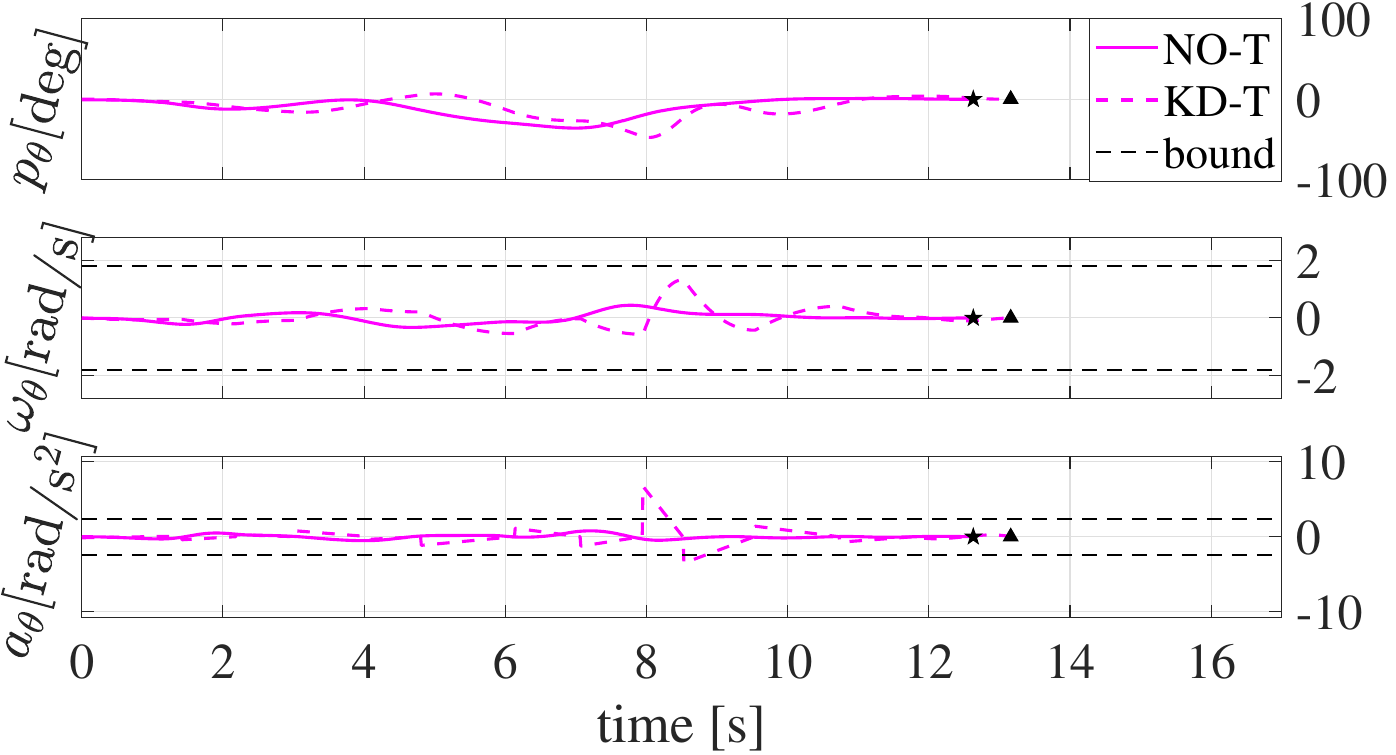}
		\end{minipage}
		\label{fig:qtraj}
	}
	\subfigure[NO-T]
	{
		\begin{minipage}[b]{.17\linewidth}
			\centering
			\includegraphics[width=3cm,frame]{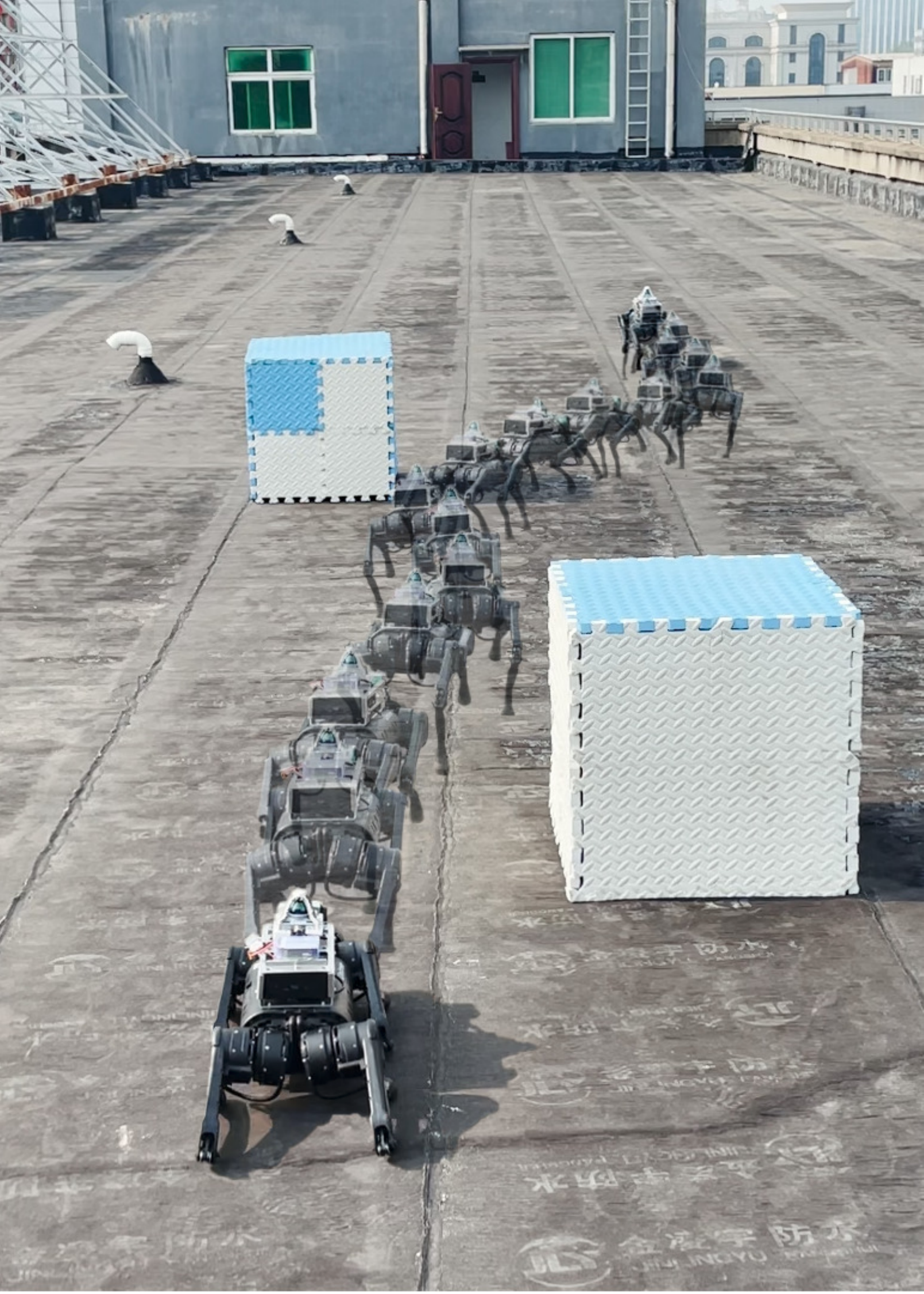}
		\end{minipage}
		\begin{minipage}[b]{.10\linewidth}
			\centering
			\includegraphics[width=2cm]{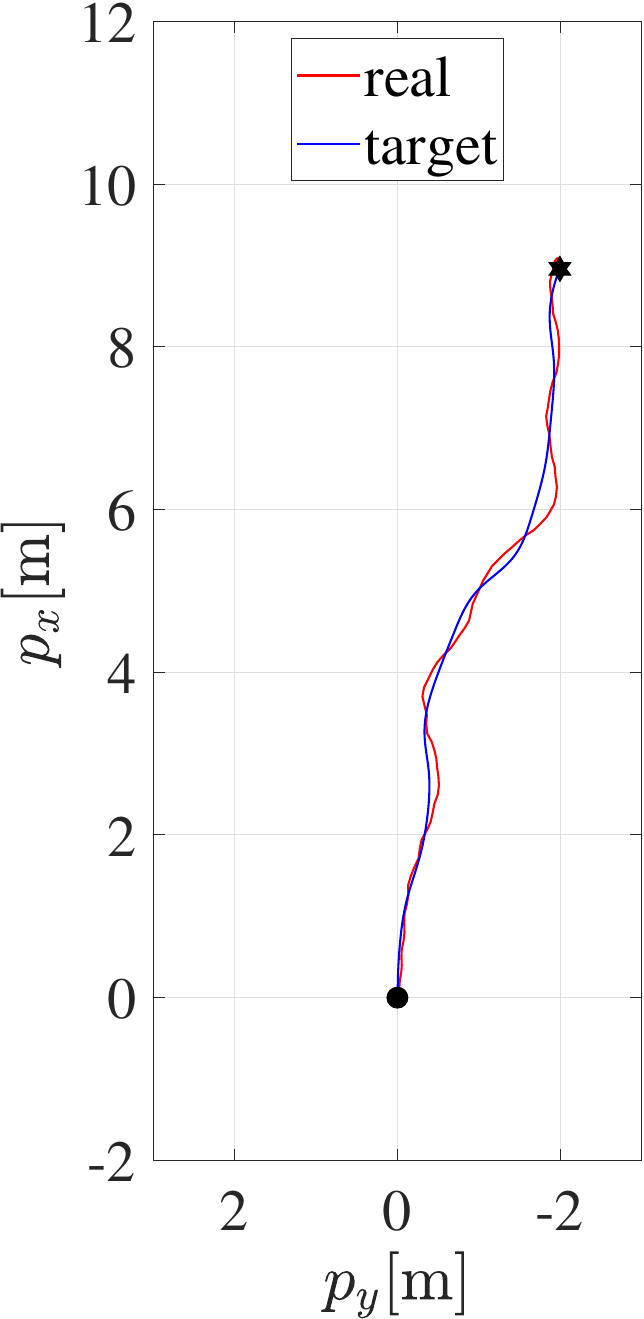}
		\end{minipage}
		\label{fig:ellipse}
	}
	\subfigure[NOY-T]{
		\begin{minipage}[b]{.18\linewidth}
			\centering
			\includegraphics[width=3cm,frame]{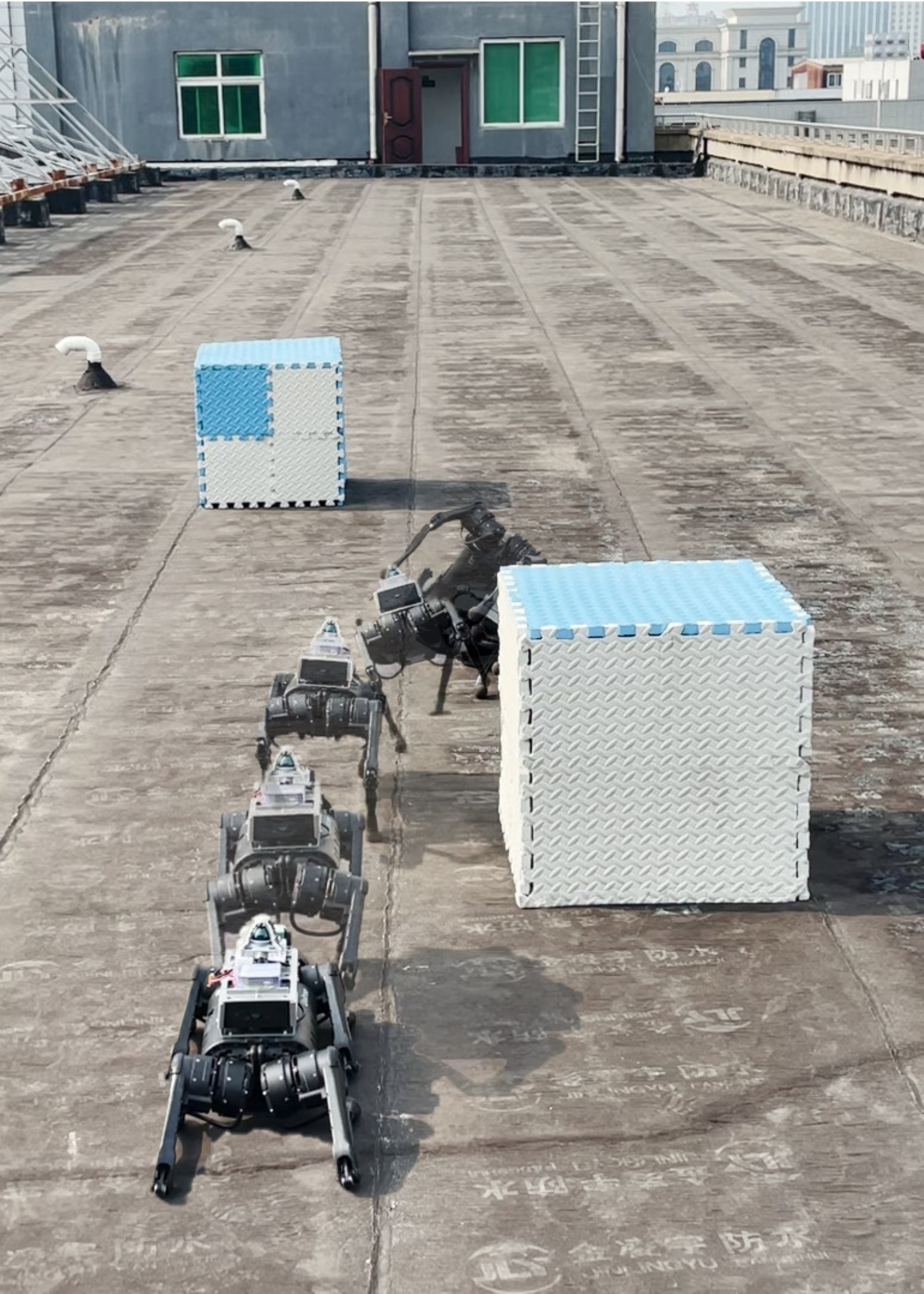}
		\end{minipage}
		\label{fig:ellipsezero}
	}
	\subfigure[KD-T]
	{
		\begin{minipage}[b]{.17\linewidth}
			\centering
			\includegraphics[width=3cm,frame]{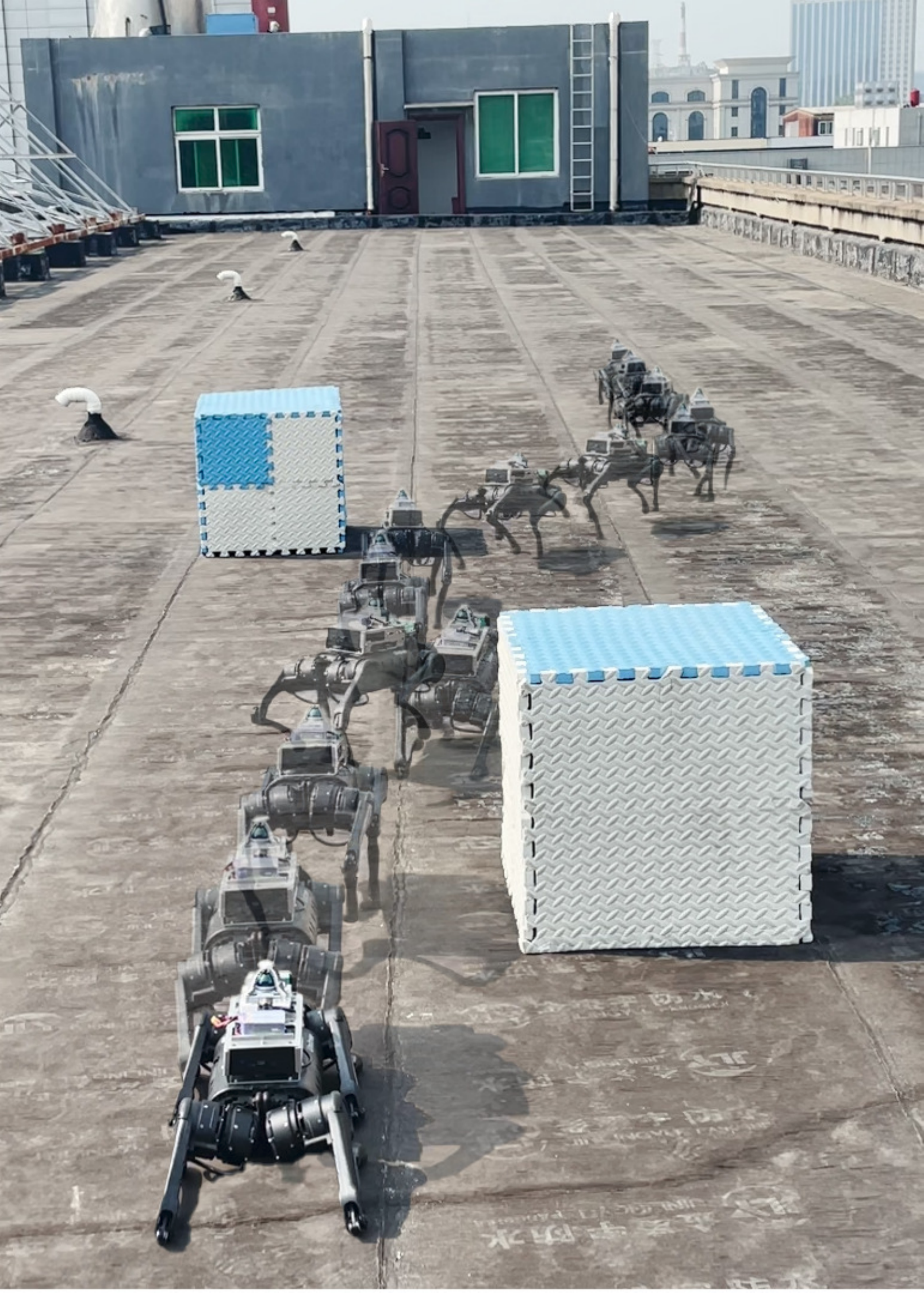}
		\end{minipage}
		\begin{minipage}[b]{.10\linewidth}
			\centering
			\includegraphics[width=2cm]{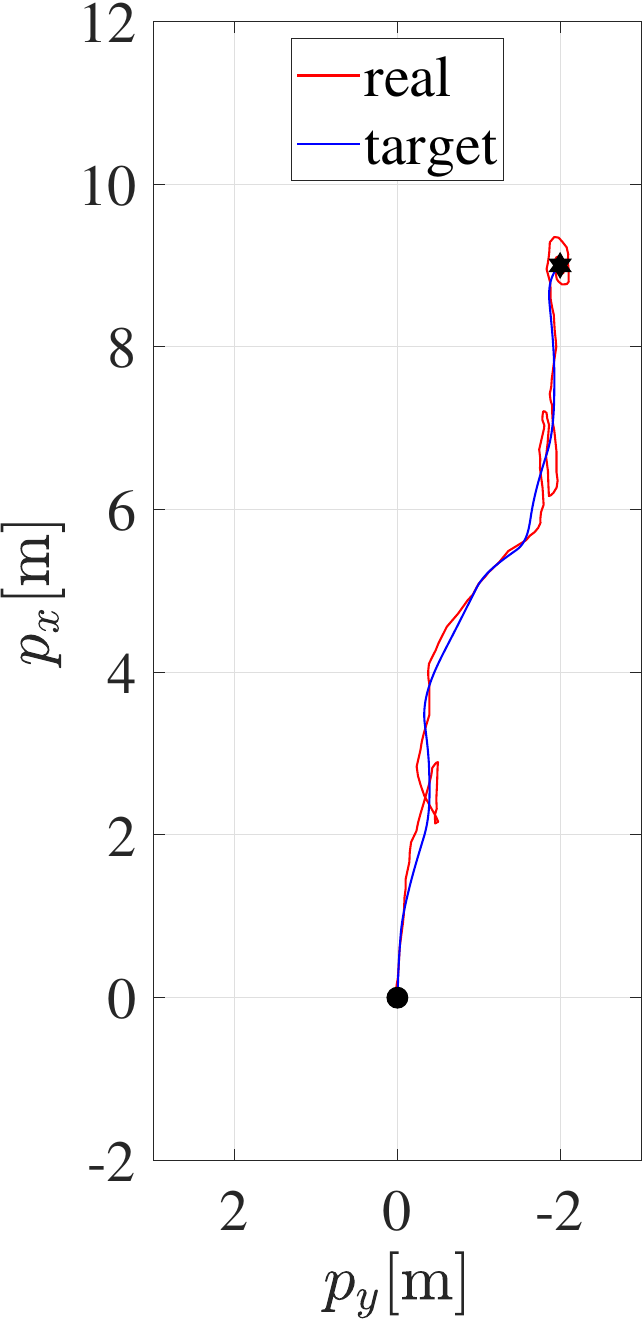}
		\end{minipage}
		\label{fig:prm}
	}
	\subfigure[KDY-T] {
		\begin{minipage}[b]{.18\linewidth}
			\centering
			\includegraphics[width=3cm,frame]{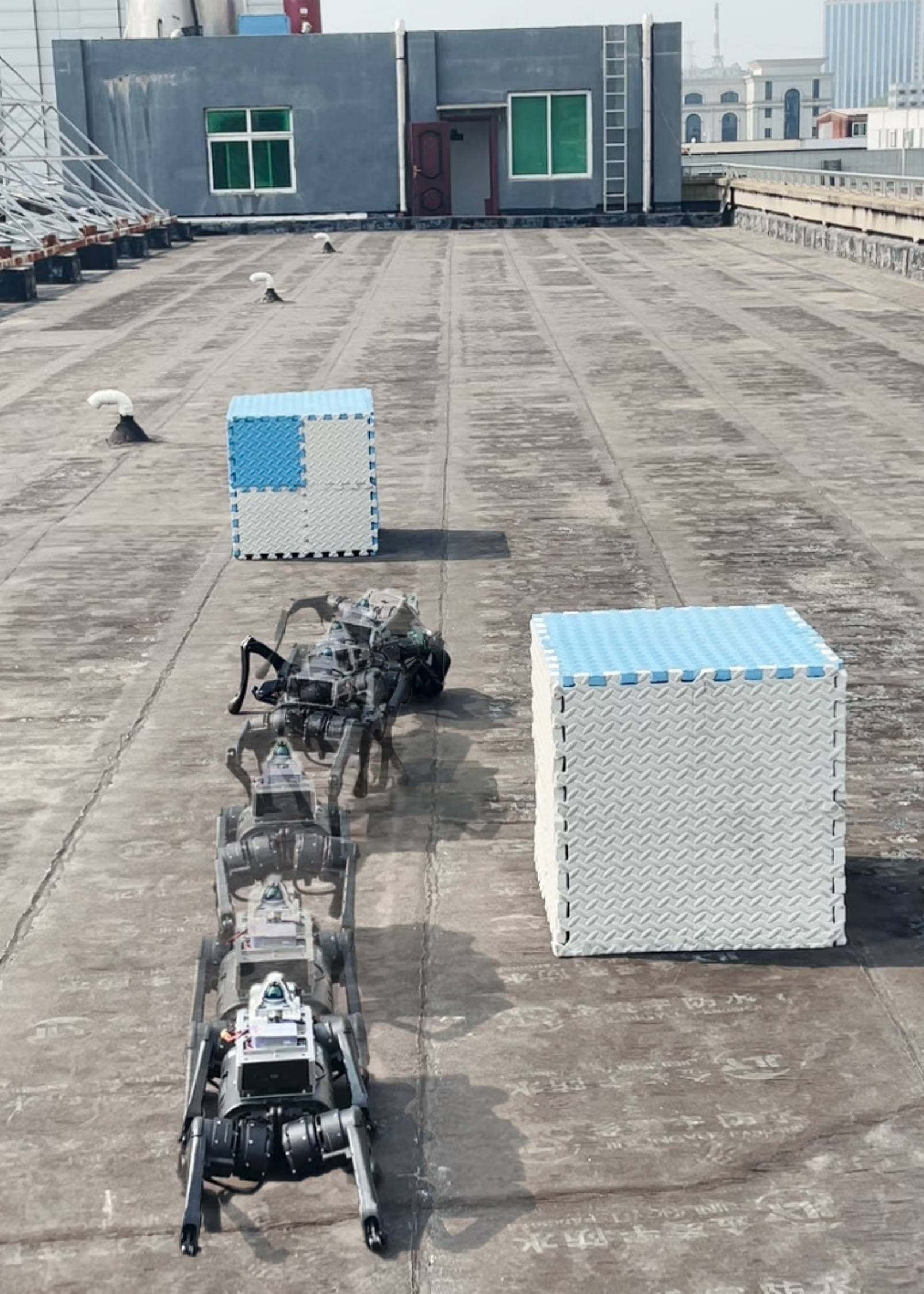}
		\end{minipage}
		\label{fig:prmzero}
	}
	\caption{(a), (b) and (c) are COM state trajectories. Both the KD-T and the NO-T exhibit continuous and smooth state transitions. Furthermore, the NO-T demonstrates a reduced time duration while satisfying the ellipse constraints. The five-pointed star serves as the destination for NO-T, while the triangle represents the endpoint for KD-T. Goal pose is [9.0,-2.0,0.0] and the ellipse velocity constraint $v_{\mathrm{mx}}$ is \SI{2.4}{m/s} and $v_{\mathrm{my}}$ is \SI{1.2}{m/s}. The $v_{\mathrm{ref}}$ is \SI{0.8}{m/s} and $\omega _{\mathrm{ref}}$ is \SI{1.2}{rad/s}, with $v_{\mathrm{max}}$ bound at $\pm$\SI{1.2}{m/s} and $\omega_{\mathrm{max}}$ bound at $\pm$\SI{1.8}{rad/s}. The $a_{\mathrm{max}}$ is $\pm$\SI{1.6}{m/s^2} and $\dot{\omega}_{\mathrm{max}}$ is $\pm$\SI{2.4}{rad/s^2}. (d) NO-T trajectory time duration is \SI{12}{s}, while (f) KD-T is \SI{21.26}{s}, (e) and (g) planning without yaw both falling over.}
	\label{fig:OMAtest}
\end{figure*}

We simulate a typical scenario in which the robot must pass through a doorway laterally to evaluate OMA for legged robots. Four experiments are conducted: optimized trajectory planning (NO-T), optimized trajectory planning without yaw angle (NOY-T), search trajectory planning (KD-T), and search trajectory planning without yaw angle (KDY-T). As shown in Fig. \ref{fig:OMAtest}, our optimization techniques effectively improve trajectory quality. It is evident from Fig. \ref{fig:prm} that the (KD-T) exhibits significant tracking errors. Nevertheless, the optimized trajectory demonstrates superior tracking performance and reaches the destination more quickly in Fig. \ref{fig:ellipse}. Notably, neglecting planning for the yaw angle renders the robot highly susceptible to falling down under the same dynamic bounds, as evidenced by Fig. \ref{fig:ellipsezero} and Fig. \ref{fig:prmzero}. Further experimental details can be found in the supplementary video.

\subsection{General Planning Experiments}
\begin{figure*}[htbp]
	\subfigure[square with obstacles]
	{
		\begin{minipage}[b]{.31\linewidth}
			\centering
			\includegraphics[width=5.5 cm,frame]{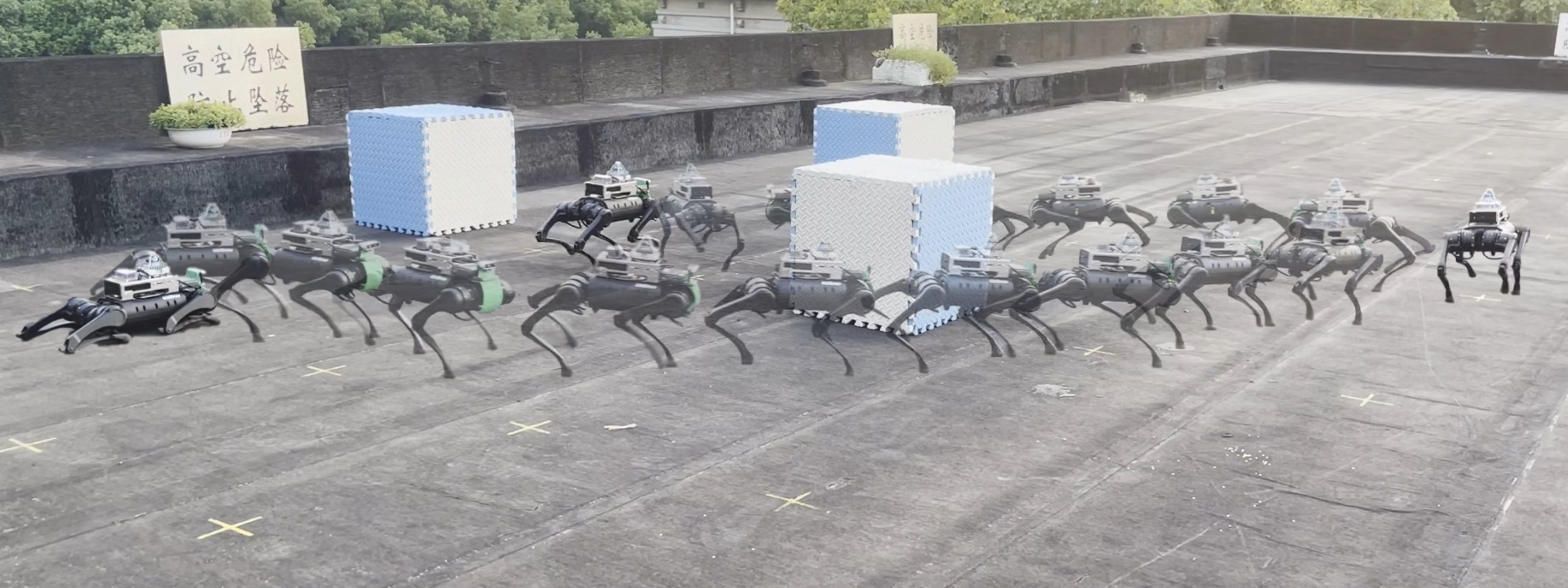} \\
			\includegraphics[width=5.5 cm,frame]{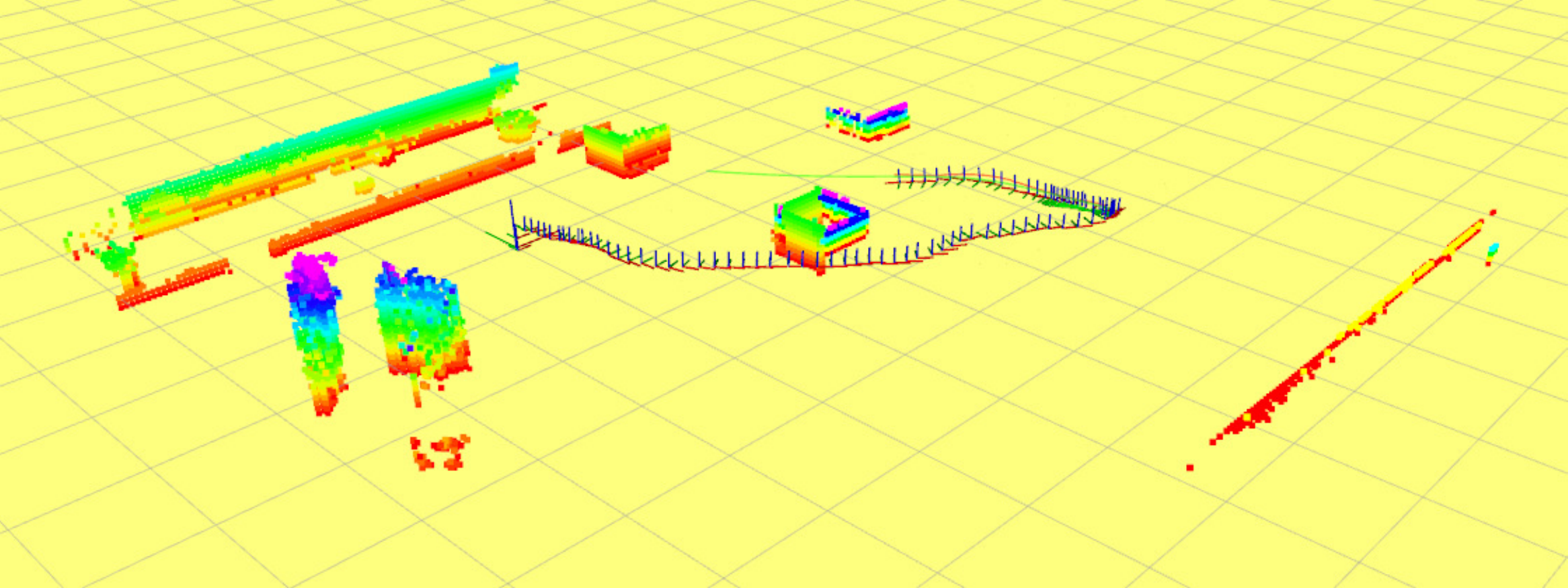}
		\end{minipage}
		\label{fig:roof}
	} 
	\subfigure[cluttered corridor]
	{
		\begin{minipage}[b]{.31\linewidth}
			\centering
			\includegraphics[width=5.5 cm,frame]{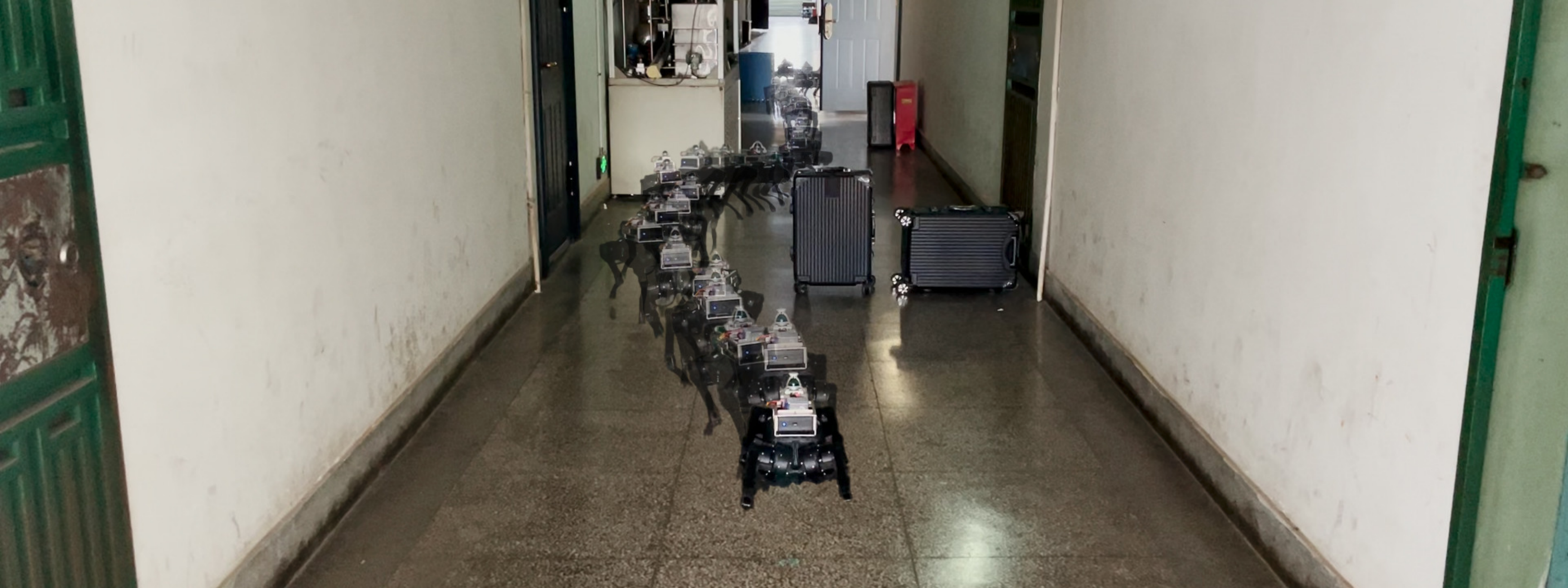} \\
			\includegraphics[width=5.5 cm,frame]{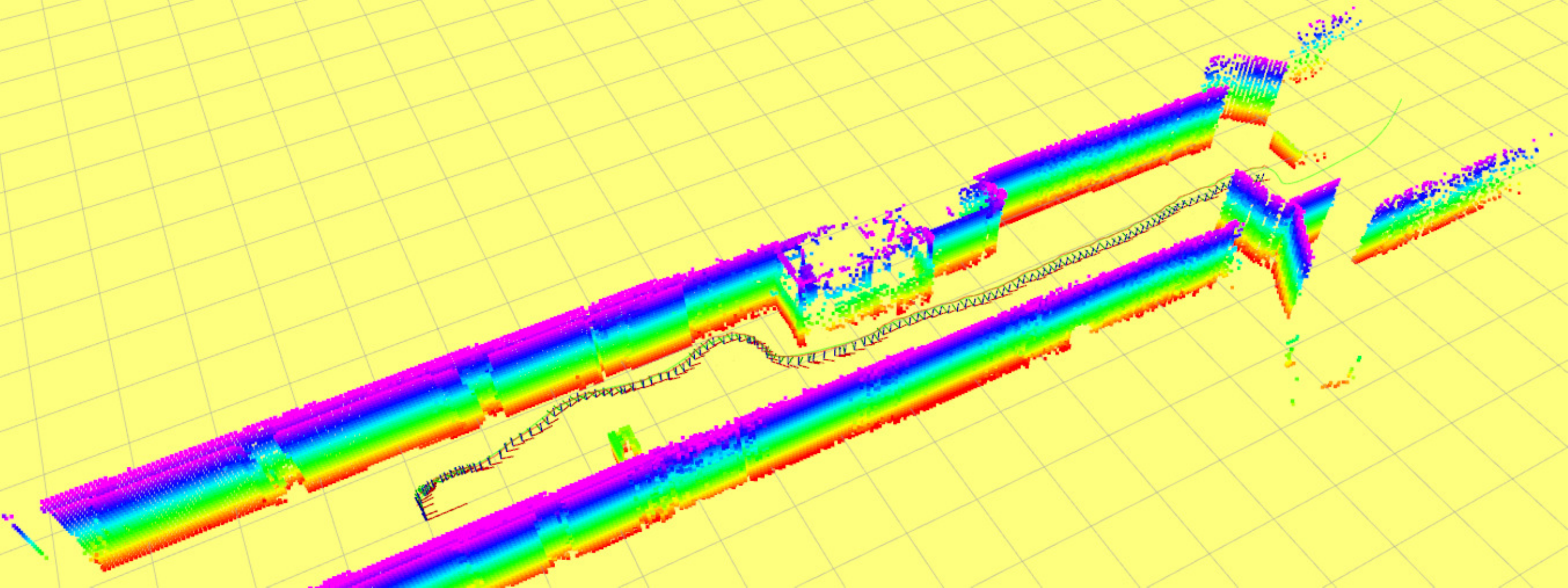}
		\end{minipage}
		\label{fig:corridor}
	}
	\subfigure[grass with trees]
	{
		\begin{minipage}[b]{.31\linewidth}
			\centering
			\includegraphics[width=5.5 cm,frame]{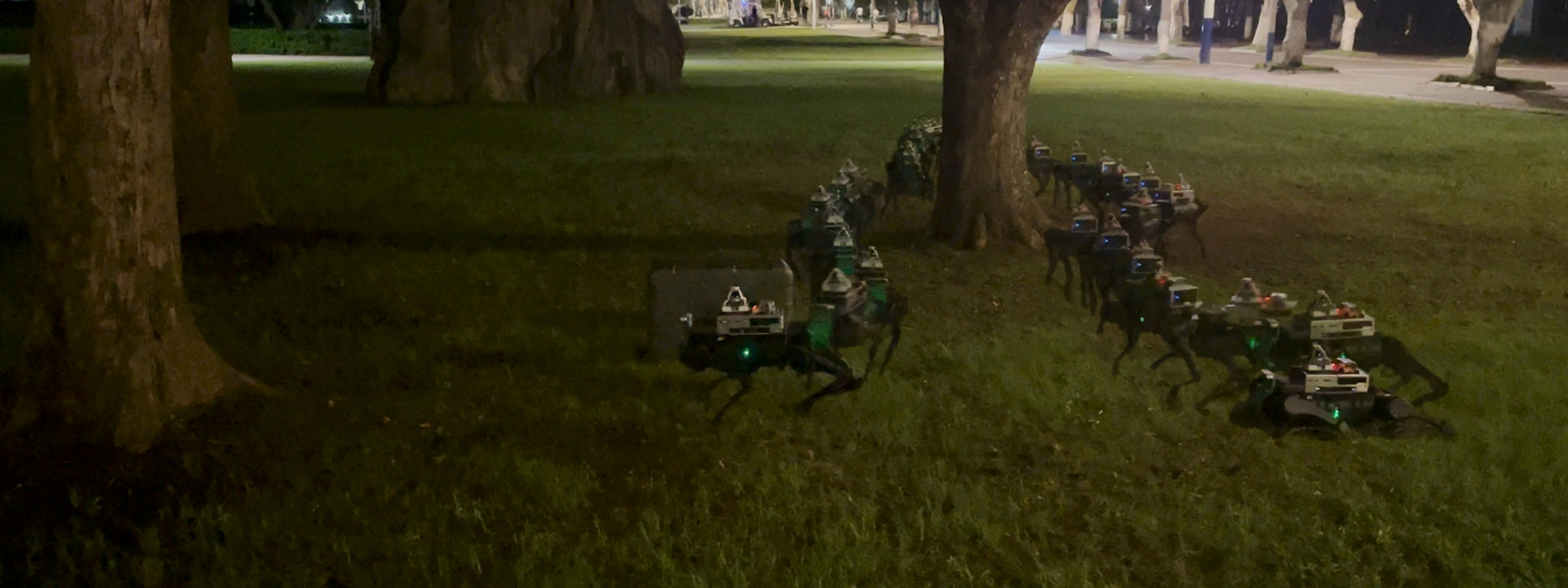} \\
			\includegraphics[width=5.5 cm,frame]{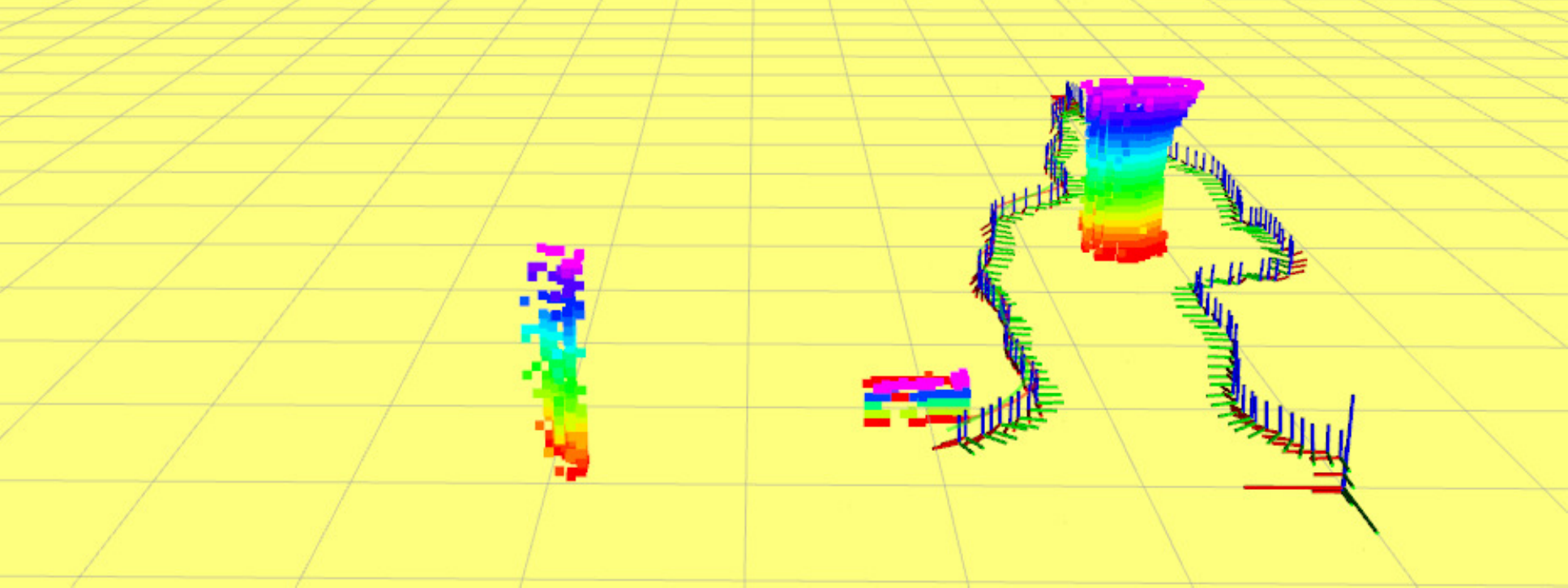}
		\end{minipage}
		\label{fig:grass}
	}
	\caption{Online planning experiments in three different scenarios. The figure above represents the actual navigation, while the figure below illustrates the information about SLAM and planning. The three-axis coordinates depict the robot's odometry.}
	\label{fig:realworldtest}
\end{figure*}

We conduct tests in three different real-world scenarios, as depicted in Fig. \ref{fig:realworldtest}. The robot successfully navigates through obstacles and efficiently reaches the desired positions and orientations, demonstrating its flexibility. Furthermore, the use of probabilistic obstacle maps enhances the robustness of the planning system against the influence of moving obstacles. The results of our real-world planning experiments are presented in Tab. \ref{tab:online}. Our planning framework proves to be adaptable to uneven grasslands while maintaining stability.
\begin{table}[htbp]
	\renewcommand{\arraystretch}{1.4}
	\centering
	\caption{General Planning Result}\label{tab:online}
	\resizebox{8.6cm}{!}{
	\begin{tabular}{ccccccc} 
	\hline
	\hline
		&	& 	\multicolumn{5}{c}{Trajectory State}  \\ 
				\cline{3-7}
	Scenario & $\boldsymbol{p_{\mathrm{goal}}}$ & $time$[\SI{}{s}] & $v_{\mathrm{ave}}$[\SI{}{m/s}] & $\omega_{\mathrm{ave}}$[\SI{}{rad/s}] & $v_{\mathrm{max}}$[\SI{}{m/s}] & $\omega_{\mathrm{max}}$[\SI{}{rad/s}] \\
	\hline
	square   & \begin{tabular}[c]{@{}c@{}} [4.0,-4.0,0.0]\\ {[}2.0,-0.5,0.0]\end{tabular} 	&19.3    &0.56	&0.50	&1.44	&-2.07  \\ 
	grass    & \begin{tabular}[c]{@{}c@{}} [0.0,-8.0,0.0]\\ {[}2.0,0.0,0.0]\end{tabular}  	&31.3    &0.56	&0.44 	&1.91	&-1.62	\\ 
	corridor & [16.0,0.0,0.0]                  											&40.7    &0.45	&0.29  	&2.87	&-2.22  \\
	\hline
	\hline
	\end{tabular}
	}
\end{table}

\section{Conclusion}\label{sec:conclusion}

In this paper, we propose a real-time hierarchical planning system for quadruped robots. We consider the anisotropy in the omnidirectional motion at different planning levels. Our navigation system enables agile motion in unknown environments and has been tested in simulations and real-world on the Unitree A1 robot platform.

In the future, we plan to expand the map representation from 2D to 2.5D or even 3D, enabling planner to tackle high-dimensional navigation challenges. Additionally, we aim to incorporate footsteps and swinging legs in local planning, while enhancing the performance of the NMPC controller and addressing uneven terrain. Our ultimate objective is to achieve more flexible and adaptable planning and control in highly complex scenarios through ongoing research efforts. 

\section*{ACKNOWLEDGMENT}
This work is supported under the National Natural Science Foundation of China under Grant 62173155 and the Program for HUST Academic Frontier Youth Team.



\newpage
\bibliographystyle{IEEEtran}   %
\bibliography{references} %
\end{document}